\documentclass[wcp]{jmlr}

\usepackage[noend]{algpseudocode}
\usepackage{algorithmicx,algorithm}
\usepackage{multicol}
\usepackage{hyperref}
\makeatletter
\let\Ginclude@graphics\@org@Ginclude@graphics 
\makeatother

\jmlrvolume{}
\jmlryear{2021}
\jmlrworkshop{Preprint}
\title[Bias-reduced Multi-step Hindsight Experience Replay]{Bias-reduced Multi-step Hindsight Experience Replay for Efficient Multi-goal Reinforcement Learning}

  \author{\Name{Rui Yang} \Email{yangrui19@mails.tsinghua.edu.cn}\\
  \Name{Jiafei Lyu} \Email{lvjf20@mails.tsinghua.edu.cn}\\
  \Name{Yu Yang} \Email{yangyu20@mails.tsinghua.edu.cn}\\
  \Name{Jiangpeng Yan} \Email{yanjp17@mails.tsinghua.edu.cn}\\
  \Name{Feng Luo} \Email{luof19@tsinghua.edu.cn}\\
  \Name{Xiu Li} \Email{li.xiu@sz.tsinghua.edu.cn}\\
  \addr Tsinghua University
  \AND
  \Name{Lanqing Li} \Email{lanqingli@tencent.com}\\
  \Name{Dijun Luo} \Email{dijunluo@tencent.com} \\
  \addr Tencent AI Lab
 }

\begin{document}

\maketitle

\begin{abstract}
Multi-goal reinforcement learning is widely applied in planning and robot manipulation. Two main challenges in multi-goal reinforcement learning are sparse rewards and sample inefficiency. Hindsight Experience Replay (HER) aims to tackle the two challenges via goal relabeling. However, HER-related works still need millions of samples and a huge computation. In this paper, we propose \emph{Multi-step Hindsight Experience Replay} (MHER), incorporating multi-step relabeled returns based on $n$-step relabeling to improve sample efficiency. Despite the advantages of $n$-step relabeling, we theoretically and experimentally prove the off-policy $n$-step bias introduced by $n$-step relabeling may lead to poor performance in many environments. To address the above issue, two bias-reduced MHER algorithms, MHER($\lambda$) and Model-based MHER (MMHER) are presented. MHER($\lambda$) exploits the $\lambda$ return while MMHER benefits from model-based value expansions. Experimental results on numerous multi-goal robotic tasks show that our solutions can successfully alleviate off-policy $n$-step bias and achieve significantly higher sample efficiency than HER and Curriculum-guided HER with little additional computation beyond HER. 
\end{abstract}

\section{Introduction}
\noindent Reinforcement learning (RL) has achieved great success in a wide range of decision-making tasks, including Atari games \citep{mnih2015human}, planning \citep{kim2009reinforcement}, and robot control \citep{kober2013reinforcement,zhang2015towards}, etc. Focusing on learning to achieve multiple goals simultaneously, multi-goal RL has also been an effective tool for sophisticated multi-objective robot control \citep{andrychowicz2017hindsight}. However, two common challenges in multi-goal RL, sparse rewards and data inefficiency limit its further application to real-world scenarios. Although designing a suitable reward function can contribute to solving these problems \citep{ng1999policy}, reward engineering itself is often challenging as it requires domain-specific expert knowledge and lots of manual adjustments. Therefore, learning from unshaped and binary rewards representing success or failure is essential for real-world RL applications.

In contrast to most of the current reinforcement learning algorithms which fail to learn from unsuccessful experiences, humans can learn from both successful and unsuccessful experiences. Inspired by such capability of humans, Hindsight Experience Replay (HER) \citep{andrychowicz2017hindsight} is proposed to learn from unsuccessful trajectories by alternating desired goals with achieved goals. Although HER has made remarkable progress to learn from binary and sparse rewards, it still requires millions of experiences for training \citep{plappert2018multi}. A few works have made efforts to improve the sample efficiency of HER \citep{zhao2018energy,fang2019curriculum}, but they don't take advantage of the intra-trajectory multi-step information, e.g., incorporating multi-step relabeled rewards to estimate future returns.

\begin{figure}[htp]
    \begin{center}
        \includegraphics[width=0.8\linewidth, trim=0 25 0 0]{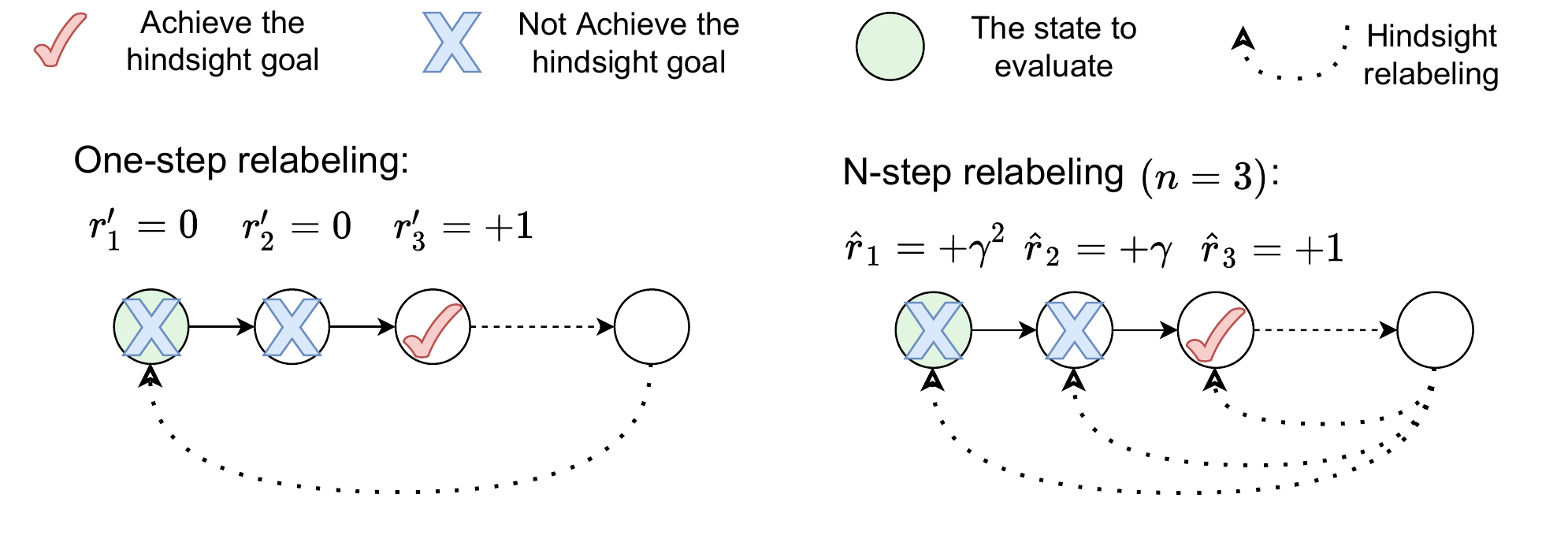}
        \caption{Diagram of $n$-step relabeling. Circles are states in a trajectory, and green circles refer to the states we evaluate. Suppose only the third state achieves the hindsight goal. One-step relabeling relabels only one transition and estimates return with the relabeled reward $r'_1$. N-step relabeling relabels $n$ (taking $n=3$ for example) consecutive transitions and combines relabeled $n$-step rewards $\hat r_1=r'_1+\gamma r'_2 + \gamma^2 r'_3$ to estimate the return, therefore it increases useful learning signals for training.}\label{fig:architecture}
    \end{center}
\vspace{-20pt}
\end{figure}

In this paper, we introduce a new framework, \emph{Multi-step Hindsight Experience Replay} (MHER), which utilizes the correlation between transitions to obtain more positive samples than HER and improves sample efficiency greatly. The core technique of MHER is $n$-step relabeling (shown in Figure \ref{fig:architecture}), which relabels $n$ consecutive transitions with achieved goals and computes relabeled $n$-step returns to estimate the value function. However, normally off-policy data in RL algorithms comes from earlier versions of the policy and is relatively on-policy, while the relabeled data in HER is completely off-policy \citep{plappert2018multi}. Experimental results and theoretical analysis consistently show that vanilla MHER may perform inconsistently in different environments due to the off-policy $n$-step bias. This problem is considered to be a promising research topic by \citep{plappert2018multi} and further leads us to propose bias-reduced MHER algorithms.

To improve the sample efficiency of HER and tackle the above issue in vanilla MHER, we propose two algorithms, MHER($\lambda$) and Model-based MHER (MMHER). Inspired by TD($\lambda$) \citep{seijen2014true}, MHER($\lambda$) combines relabeled $n$-step returns with exponentially decayed weights parameterized by $\lambda$, making a trade-off between bias and reward signals. Unlike MHER($\lambda$), MMHER alleviates off-policy bias by generating on-policy returns with a trained dynamics model. We conduct detailed experiments on eight challenging simulated robot environments \citep{plappert2018multi,yu2020meta}, including \emph{Sawyer}, \emph{Fetch} and \emph{Hand} environments. Experimental results \footnote{Anonymous code is available at \href{https://anonymous.4open.science/r/c2265620-8572-4375-8a92-251cc35c4b2e}{https://anonymous.4open.science/r/c2265620-8572-4375-8a92-251cc35c4b2e}} demonstrate that both of our two algorithms can alleviate off-policy $n$-step bias and outperform HER and Curriculum-guided HER (CHER) \citep{fang2019curriculum} in sample efficiency. To the best of our knowledge, our work is the first to successfully incorporate multi-step information to solve sparse-reward multi-goal RL problems.

In conclusion, our contributions can be summarized as follows:
\begin{itemize}
    \item[1)]We present the framework of MHER based on $n$-step relabeling to incorporate multi-step relabeled rewards into HER;
    \item[2)]We analyze the off-policy $n$-step bias introduced by MHER theoretically and experimentally;
    \item[3)]We propose two simple but effective bias-reduced MHER algorithms, MHER($\lambda$) and MMHER, both of them successfully mitigate the off-policy $n$-step bias and achieve significantly higher sample efficiency than HER and CHER. 
\end{itemize}

\section{Related Work}
In reinforcement learning, multi-step methods \citep{sutton1988learning,watkins1989learning} have been studied for a long history since Monte Carlo (MC) and Temporal Difference (TD). Q($\lambda$)-learning \citep{peng1994incremental} combines Q-learning and TD($\lambda$) for faster learning and alleviating the non-Markovian effect. Two well-known works, Rainbow \citep{hessel2018rainbow} and D4PG \citep{barth2018distributed}, utilize $n$-step returns to improve the performance of DQN \citep{mnih2015human} and DDPG \citep{lillicrap2015continuous}. However, $n$-step returns in off-policy algorithms are biased due to policy mismatch. One bias-free solution is \emph{off-policy correction} using Importance Sampling (IS) \citep{precup2000eligibility}. Retrace($\lambda$) \citep{munos2016safe} is introduced to reduce the variance in IS with guaranteed efficiency. Overall, multi-step methods provide a forward view and contribute to a faster propagating of value estimation.

In contrast to multi-step methods, HER \citep{andrychowicz2017hindsight} provides a hindsight view in completed trajectories. Through hindsight relabeling with achieved goals, HER hugely boosts the sample efficiency in complex environments with sparse and binary rewards. A series of algorithms have been proposed to further improve the performance of HER. Energy-Based Prioritization \citep{zhao2018energy} prioritizes experiences based on the trajectory energy for more efficient usage of collected data.
% G-HER \citep{bai2019guided} utilizes a conditional recurrent neural network to generate guided goal sequences for HER.
Curriculum-guided HER \citep{fang2019curriculum} adaptively makes an exploration-exploitation trade-off to select hindsight goals. Maximum Entropy-based Prioritization \citep{zhao2019maximum} introduces a weighted entropy to encourage agents to achieve more diverse goals. Concentrating on exploiting the information between transitions, our work is orthogonal to these algorithms. 

In general, model-based RL algorithms have the advantage of higher sample efficiency over model-free algorithms \citep{nagabandi2018neural}. Dyna \citep{sutton1991dyna} utilizes a trained model to generate virtual transitions for accelerating learning value functions. However, model-based methods are limited by model bias, as previous work \citep{janner2019trust} addressed. Our model-based algorithm is closely related to Model-Based Value Expansion (MVE) \citep{feinberg2018model}, which applies model-based value expansion to enhance the estimation of expected returns. The differences are: 1) our model-based return is under the multi-goal setting, and 2) we utilize a compound form of MVE and hindsight relabeling to jointly exploit MVE and hindsight knowledge (see Section \ref{sec:MMHER}).

\section{Preliminaries}
\subsection{Reinforcement Learning}
Reinforcement learning (RL) solves the problem of how an agent acts to maximize cumulative rewards obtained from an environment. Generally, the RL problem is formalized as a \emph{Markov Decision Process} (MDP), which consists of five elements, state space $\mathcal{S}$, action space $\mathcal{A}$, reward function $r: \mathcal{S} \times \mathcal{A} \rightarrow \mathbb{R}$, transition probability distribution $p(s'|s,a)$, and discount factor $\gamma \in [0,1]$. The agent learns a policy $\pi: \mathcal{S} \rightarrow \mathcal{A}$ to maximize expected cumulative rewards $E[\sum_{t=0}^{\infty} \gamma^t r_t]$. The Q-function $Q:\mathcal{S} \times \mathcal{A} \rightarrow \mathbb{R}$ is defined as the expected return starting from the state-action pair $(s_t,a_t)$: 
\begin{equation*}
    Q(s_t,a_t) = E\big[\sum_{k=0}^{\infty} \gamma^k r_{t+k}|s_t, a_t\big]
\end{equation*}

\subsection{Deep Deterministic Policy Gradient (DDPG)}
DDPG \citep{lillicrap2015continuous} is an off-policy algorithm for continuous control. An actor-critic structure is applied in DDPG, where the actor serves as the policy $\pi$ and the critic approximates the Q-function. Denote the replay buffer as $B$. The actor is updated with gradient descent on the loss: 
\begin{equation*}
    L_{actor}=-E_{s_t\sim B}[Q(s_t,\pi(s_t))].
\end{equation*}
The critic is updated to minimize the TD error:
\begin{equation*}
    {L}_{critic}=E_{(s_t,a_t,s_{t+1}) \sim B}[(y_t-Q(s_t, a_t))^2],
\end{equation*}
where $y_t=r_t +\gamma Q(s_{t+1}, \pi (s_{t+1}))$.

In D4PG \citep{barth2018distributed}, $n$-step returns are utilized to enhance the approximation of the value function, where $n$-step target $y^{(n)}_t$ is defined as:
\begin{equation}
\label{eq:d4pgtarget}
    y^{(n)}_t = \sum_{k=0}^{n-1} \gamma ^k r_{t+k} +  \gamma ^ n Q(s_{t+n},
    \pi(s_{t+n})) .
\end{equation}

\subsection{Hindsight Experience Replay (HER)}
\label{sec:her}
HER \citep{andrychowicz2017hindsight} is proposed to tackle sparse reward problems in multi-goal RL. Following \emph{Universal Value Function Approximators} (UVFA) \citep{schaul2015universal}, HER considers goal-conditioned policy $\pi:\mathcal{S} \times \mathcal{G} \rightarrow \mathcal{A}$ and value function $Q:\mathcal{S} \times \mathcal{A} \times \mathcal{G} \rightarrow \mathbb{R}$. The sparse reward function is also conditioned by goals:
\begin{equation}
\label{equ:rewardfunction}
    r(s_t, a_t,g)=\begin{cases}0, & \left|\left|\phi(s_{t})-g\right|\right|^2_2 < \text{threshold} \cr -1, &\text{otherwise}\end{cases} ,
\end{equation}
where $\phi:\mathcal{S} \rightarrow \mathcal{G}$ maps states to goals. The key technique of HER is hindsight relabeling, which relabels transition $(s_t, a_t, s_{t+1}, r_t, g)$ with achieved goal in the same trajectory $g'=\phi(s_{t+k}), k\geq 0$ and new reward $r'_{t}=r(s_t, a_t, g')$ according to Eq. (\ref{equ:rewardfunction}). After hindsight relabeling, HER augments training data with relabeled transitions $(s_t, a_t, r'_t, s_{t+1}, g')$ and increases the proportion of success trails, thereby alleviating the sparse reward problem and significantly improving sample efficiency.

\begin{figure}[tb]
\begin{center}
    \includegraphics[width=1\linewidth,trim=0 50 0 0]{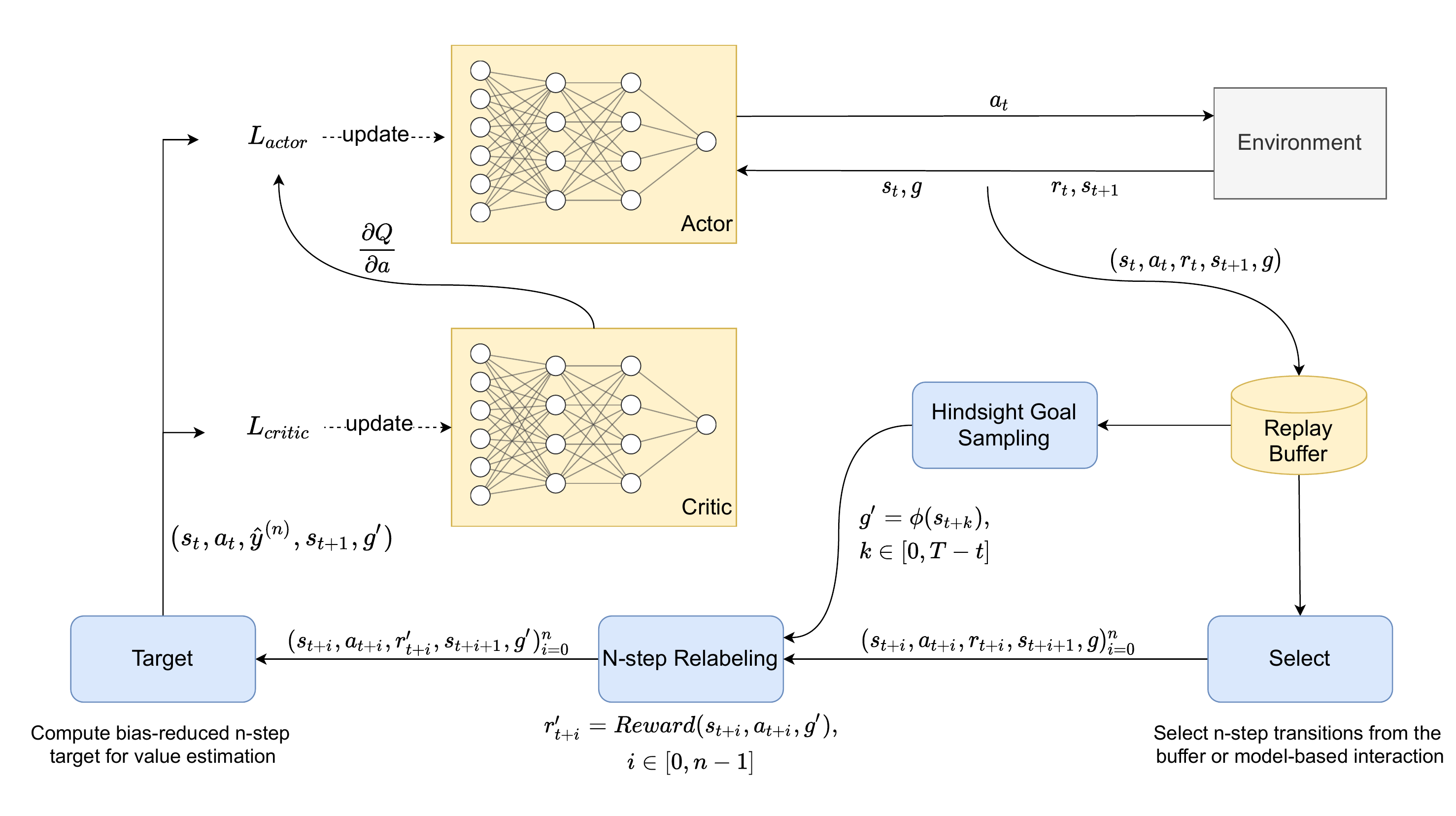}
    \caption{The Overall structure of MHER. Yellow modules are from DDPG, and blue modules belong to MHER. Major differences between HER and MHER are the blue modules in the bottom row. MHER selects $n$-step transitions using \emph{Select} function and relabels these transitions using \emph{N-step Relabeling}. After that we compute target with \emph{Target} function to update the value function.}
    \label{fig:mher_framework}
\end{center}
\vspace{-20pt}
\end{figure}

\begin{algorithm2e}[htb]
	\caption{MHER Framework \\
	{\bf Given:} an off-policy RL algorithm $A$, a replay buffer $B$, a reward function $Reward$, the number of sample goals $k$, step number $n$, a target function $Target$, a selection function $Select$.
	\label{framwork}} 
	
	Initialize policy $\pi$ and value function $Q$ \;
		\For {episode = $1,2,\ldots,M$}{
		    Sample a desired goal $g$\;
		    Collect a $T$-step trajectory with $\pi$ and save to $B$\;
		    Sample a batch $b=\{(s_t,a_t,s_{t+1},r_t,g)_m\}^N_{m=1} \sim B$\;
		    \For {$m=1,2, \ldots, N$}{
		        $(s_t,a_t,s_{t+1},r_t,g) \leftarrow b_m$ and set $g' = g$\;
		        \If {$random() < \frac{k}{1+k}$}{
		            Sample future achieved goal $g'$
		        }
		        Select $n$ transitions using $Select(*args)$\;
		        Relabel $n$ transitions with $g'$\;
		        Recompute each $r'_{t+i}$ with $Reward(* args)$\;
		        Compute target $\hat y^{(n)}=Target(*args)$\;
		        $b_m \leftarrow (s_t, a_t, \hat y^{(n)}, s_{t+1}, g')$
		    }
		    Update $Q$ with $b$ by minimizing the loss: \\
		    \centerline{$\frac{1}{N} \Sigma (\hat y^{(n)} -Q(s_t,a_t,g'))^2$} 
		    Update $\pi$ with $b$ according to $A$ \;
		}
\end{algorithm2e}

\section{Multi-step Hindsight Experience Replay}
In this section, we first introduce $n$-step relabeling and the MHER framework. Then we show a motivating example where vanilla MHER performs poorly. For demonstration purposes, all formulations are conducted in the DDPG+HER framework \citep{andrychowicz2017hindsight,fang2019curriculum}.
% We also explain why previous work Rainbow and D4PG don't need off-policy correction through statistical analysis.

\subsection{$N$-step Relabeling}
In \emph{$n$-step relabeling}, multiple transitions rather than just a single transition are relabeled, therefore more information between relabeled transitions can be exploited. Given $n$ consecutive transitions $\{(s_{t+i}, a_{t+i},r_{t+i},s_{t+1+i},g) \}_{i=0}^{n-1}$ in a collected trajectory of length $T$, we alternate goals and rewards with $g'=\phi(s_{t+k}), k\in [0,T-t]$ and $r'_{t+i}=r(s_{t+i}, a_{t+i}, g'),i\in[0,n-1]$, and then obtain relabeled transitions $\{(s_{t+i}, a_{t+i},r'_{t+i},s_{t+1+i},g') \}_{i=0}^{n-1}$.
After relabeling, we can compute relabeled $n$-step returns for value estimation:
\begin{equation}
\label{eq:nsteptarget}
y_{t}^{(n)} = \sum^{n-1}_{i=0}\gamma^i r'_{t+i} + \gamma^n Q(s_{t+n}, \pi(s_{t+n}, g'), g'),
\end{equation}
where goal $g'$ is included in action-value function $Q$ and policy $\pi$ following UVFA \citep{schaul2015universal}.

Consider a situation where only one state in the trajectory achieves the hindsight goal, as shown in Figure \ref{fig:architecture}. One-step relabeling only holds a single non-negative sample for this trajectory while $n$-step relabeling possesses $n$ non-negative samples. Therefore, $n$-step relabeling remarkably contributes to solving sparse reward problems, for which non-negative learning signals are very crucial.

\subsection{MHER Framework}
The overall framework of MHER is presented in Algorithm \ref{framwork} and Figure \ref{fig:mher_framework}. After collecting a $T$-step trajectory into the replay buffer, we randomly sample a minibatch $b$ from the buffer. For every transition in $b$ we sample a future goal $g'$ according to a certain probability $\frac{k}{1+k}$, where $k$ represents the ratio of the relabeled data to the original data. Next, for each transition we select $n$ consecutive transitions with the $Select$ function. Then, we perform $n$-step relabeling and compute relabeled $n$-step target $\hat y^{(n)}$ using the $Target$ function. Finally, the Q-function is updated with target $\hat y^{(n)}$ and the policy is trained by any off-policy RL algorithm such as DDPG and SAC \citep{haarnoja2018soft}. For vanilla MHER, the $Target$ function is Eq.(\ref{eq:nsteptarget}) and the $Select$ function outputs $n$ consecutive transitions in the same trajectory: $\{(s_{t+i}, a_{t+i},r_{t+i},s_{t+1+i},g) \}_{i=0}^{n-1}, n=min\{n, T-t\}$.

\begin{figure*}
\centering
\subfigure[]{
    \begin{minipage}[t]{0.45\linewidth}
        \centering
        \includegraphics[width=0.8\linewidth]{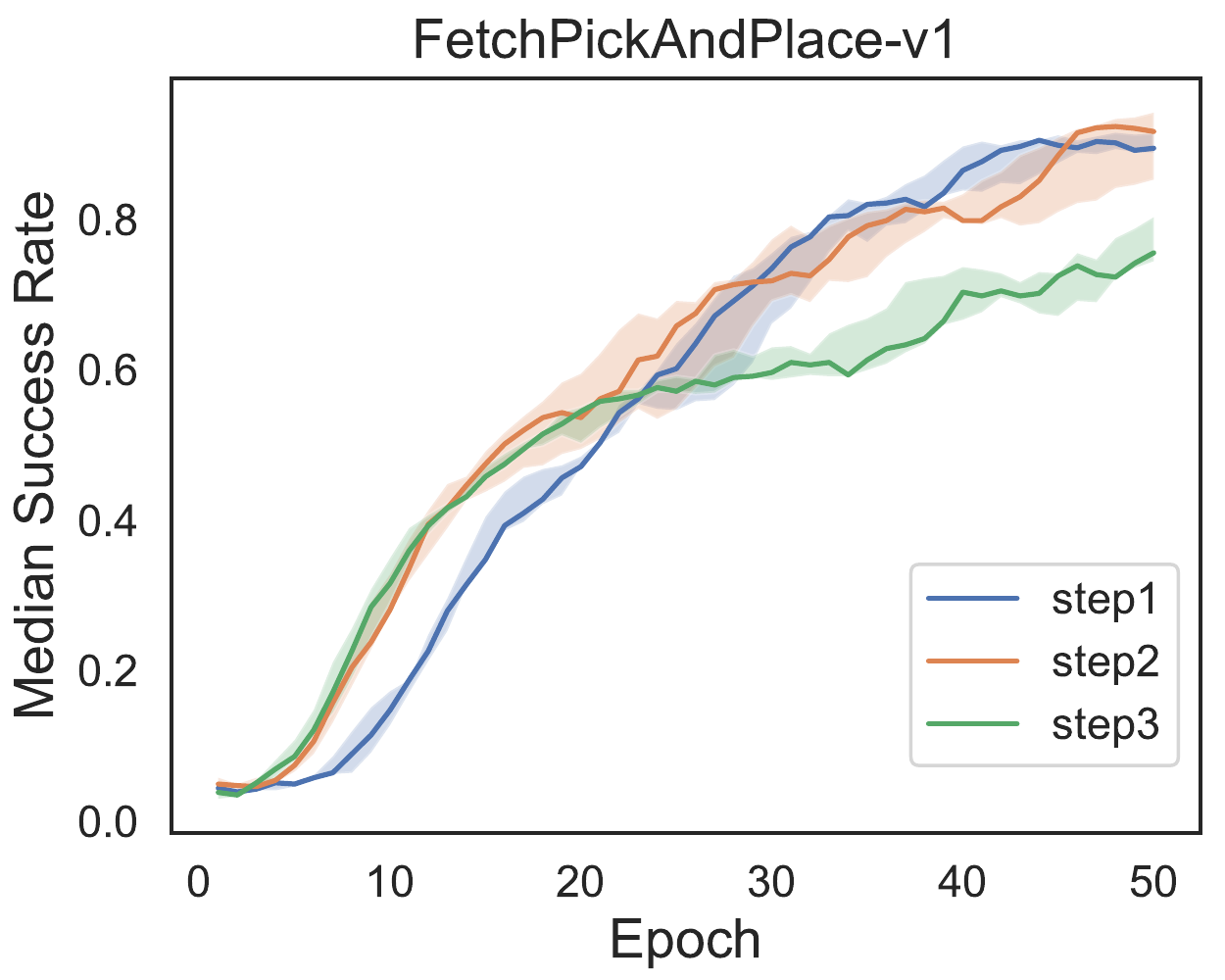}\\
    \end{minipage}%
}%
\subfigure[]{
    \begin{minipage}[t]{0.45\linewidth}
        \centering
        \includegraphics[width=0.8\linewidth]{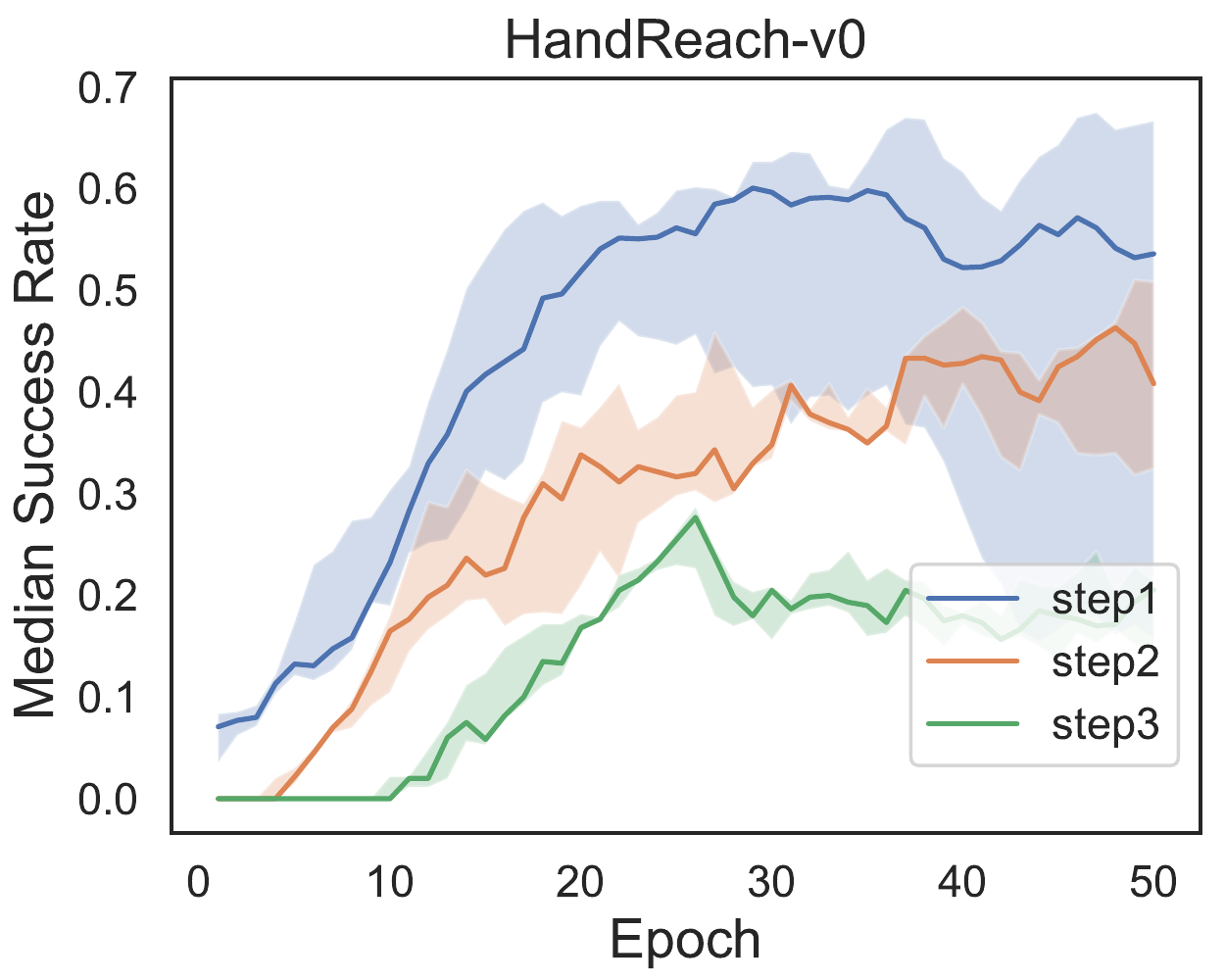}\\
    \end{minipage}%
}%
\centering
\caption{Performance of vanilla MHER in simulated robotics environments. There is a clear performance decreases when steps increase.}
\label{fig:nstepher}
\vspace{-15pt}
\end{figure*}

% \begin{figure*}
%     \centering
%     \includegraphics[scale=0.4, trim = 0 0 0 0,clip]{nstep-her.pdf}
%     \caption{Performance of vanilla MHER in simulated robotics environments. There is a clear performance decreases when steps increase.}
%     \label{fig:nstepher}
% \end{figure*}

\subsection{A Motivating Example of Bias-reduced MHER}
We compare vanilla MHER with HER ($n=1$) in simulated robotics environments and results are shown in Figure \ref{fig:nstepher}. There is a clear performance decrease when $n$ is large in both Fetch and Hand environments. In addition, the performance decrease in HandReach is more significant than that in FetchPickAndPlace. The primary reason behind this is the off-policy $n$-step bias, which we will discuss in the following section.

% \begin{figure*}
% \centering
% \subfigure[]{
%     \begin{minipage}[t]{0.4\textwidth}
%         % \centering
%         \includegraphics[width=5.2cm ]{AuthorKit21ICAPS/LaTeX/bias.pdf}\\
%     \end{minipage}%
% }%
% \subfigure[]{
%     \begin{minipage}[t]{0.4\textwidth}
%         % \centering
%         \includegraphics[width=5cm]{AuthorKit21ICAPS/LaTeX/reward.pdf}\\
%     \end{minipage}%
% }%
% % \subfigure[]{
% %     \begin{minipage}[t]{0.33\textwidth}
% %         % \centering
% %         \includegraphics[ width=6cm ]{AuthorKit21ICAPS/LaTeX/mean-reward.pdf}\\
% %     \end{minipage}%
% % }%
% \centering
% \caption{(a) Off-policy $n$-step bias training using vanilla MHER in FetchSlide and HandReach environments. (b) Average rewards training using HER in Robotics environments.}
% % \hspace{1cm}
% \label{fig:rewardsandbias}
% \end{figure*}
% % (c) Average rewards training using DDPG or DQN (without relabeling) in typical Robotics, MuJoCo and Atari environments

\section{Off-policy $N$-step Bias}
\label{sec:bias}
The relabeled $n$-step returns in MHER introduce bias due to the discrepancy of the data collecting policy and the target policy. Such bias is the main reason for MHER's poor performance in Figure \ref{fig:nstepher}. 
In this section, we study the off-policy $n$-step bias through theoretical and empirical analysis. For brevity, we omit goals $g$ as it can be included as part of states.
% , which is environment-specific and associated with the average reward.  
% We provide a detailed analysis on the form of off-policy $n$-step bias in Proposition \ref{propositionbias}.

\begin{proposition}
\label{propositionbias}
Given $n$-step transitions under data collecting policy $\mu$: $\{(s_t, a_t, r_t, s_{t+1})\}_{t=0}^{n-1}$, target policy $\pi$, discount factor $\gamma$, action value function $Q$. Policies $\mu$ and $\pi$ are deterministic. The off-policy $n$-step bias $\mathcal{B}_t^{(n)}$ at time step $t$ has the following formula:
\begin{equation}
\label{eq:off-policybias}
    \mathcal{B}_t^{(n)} = \sum_{i=1}^{n-1} \gamma^i [Q(s_{t+i},\pi(s_{t+i}))-Q(s_{t+i},a_{t+i})].
\end{equation}
\end{proposition}
The proof is provided in Appendix \ref{ap:proof1} in the supplementary file. Although the sign of the $i$-th term in Eq. (\ref{eq:off-policybias}) is uncertain, each term has a higher probability of being non-negative as the policy $\pi$ is learned to maximize $E[Q(s,\pi(s))]$. 
For further analysis, we define the average off-policy $n$-step bias and the absolute average reward in the replay buffer as:
\begin{equation}
\label{eq:averagebias}
     \mathcal{B}^{(n)}=E_{(s_{t+i},a_{t+i},s_{t+i+1})_{i=0}^{n-1}\sim B}\big[ \mathcal{B}_t^{(n)} \big]
\end{equation}
\begin{equation}
    \label{eq:abs_avg_reward}
    \bar r_{abs} = |E_{r_t\sim B} [r_t]|
\end{equation}
where $B$ refers to the replay buffer, and $\bar r_{abs}$ reflects the difficulty of the task to a certain extent when using the sparse reward function. 

\subsection{Intuitive Analysis of $\mathcal{B}^{(n)}$} 
\label{sec:intutive}
From Eq. (\ref{eq:off-policybias}) and Eq. (\ref{eq:averagebias}), we can conclude that there are three key factors affecting $\mathcal{B}^{(n)}$, step number $n$,  policy $\pi$, and value function $Q$: 
\begin{itemize}
    \item [1)] $\mathcal{B}^{(n)}$ equals zero when $n=1$ and accumulates as $n$ increases;
    \item [2)] $\mathcal{B}^{(n)}$ equals zero when $\pi=\mu$ ($\mu$ is the data collecting policy), otherwise accumulates as the shift of $\pi$ and $\mu$ increases;
    \item [3)] $\mathcal{B}^{(n)}$ increases as the magnitude of the gradient of $Q$ (related to average reward) increases.
\end{itemize}
Note that in MHER setting, the data collecting policy $\mu$ is the policy generating the relabeled data and differs from early versions of the agent's policy \citep{plappert2018multi}. Therefore, the second factor is more difficult to control than previous works \citep{hessel2018rainbow,barth2018distributed}. In Proposition \ref{propsitionaveragebias}, we give an upper bound on $\mathcal{B}^{(n)}$ to show the connection between $\mathcal{B}^{(n)}$ and the absolute average reward $\bar r_{abs}$ in Eq. (\ref{eq:abs_avg_reward}).

\begin{proposition}
\label{propsitionaveragebias}
Denote action value function under current policy $Q(s,\pi(s))$ as $V^{\pi}(s)$. The reward function is defined in Eq. (\ref{equ:rewardfunction}). Assuming $V^{\pi}(s)$ is Lipschitz continuous with constant $L$ for $\forall s,s' \in S$, the average off-policy $n$-step bias can be bounded as follows:
\begin{equation}
\label{eq:biasbound}
\begin{aligned}
    \mathcal{B}^{(n)}\leq \gamma(n-1)\big[ |E_{r_t\sim B} [r_t]| + \gamma L E_{(s_t,s_{t+1})\sim B} \|s_t - s_{t+1}\|\big]
\end{aligned}
\end{equation}
\end{proposition} 
The proof and the tightness analysis of the bound are in Appendix \ref{ap:proof2}. Proposition \ref{propsitionaveragebias} evidently indicates that the absolute average reward has an impact on the average off-policy $n$-step bias $B^{(n)}$. Furthermore, $B^{(n)}$ is environment-specific as the absolute average reward varies with different environmental difficulties, i.e., tasks with larger absolute average reward may obtain larger off-policy $n$-step bias.

%  In addition, the i-th term in Eq. (\ref{eq:off-policybias}) can be bounded by the difference between the maximum reward and minimum reward. Using reward function in Eq. (\ref{equ:rewardfunction}), the average bias is also related to the average absolute reward. 
% Therefore, we define reward density as $E_{r\sim B}[|r|]$ for further analysis, where $B$ refers to the replay buffer.
\subsection{Empirical Analysis of $\mathcal{B}^{(n)}$}
\begin{figure}
\subfigure[]{
    \begin{minipage}[t]{0.45\linewidth}
    \centering
    \includegraphics[width=0.85\linewidth]{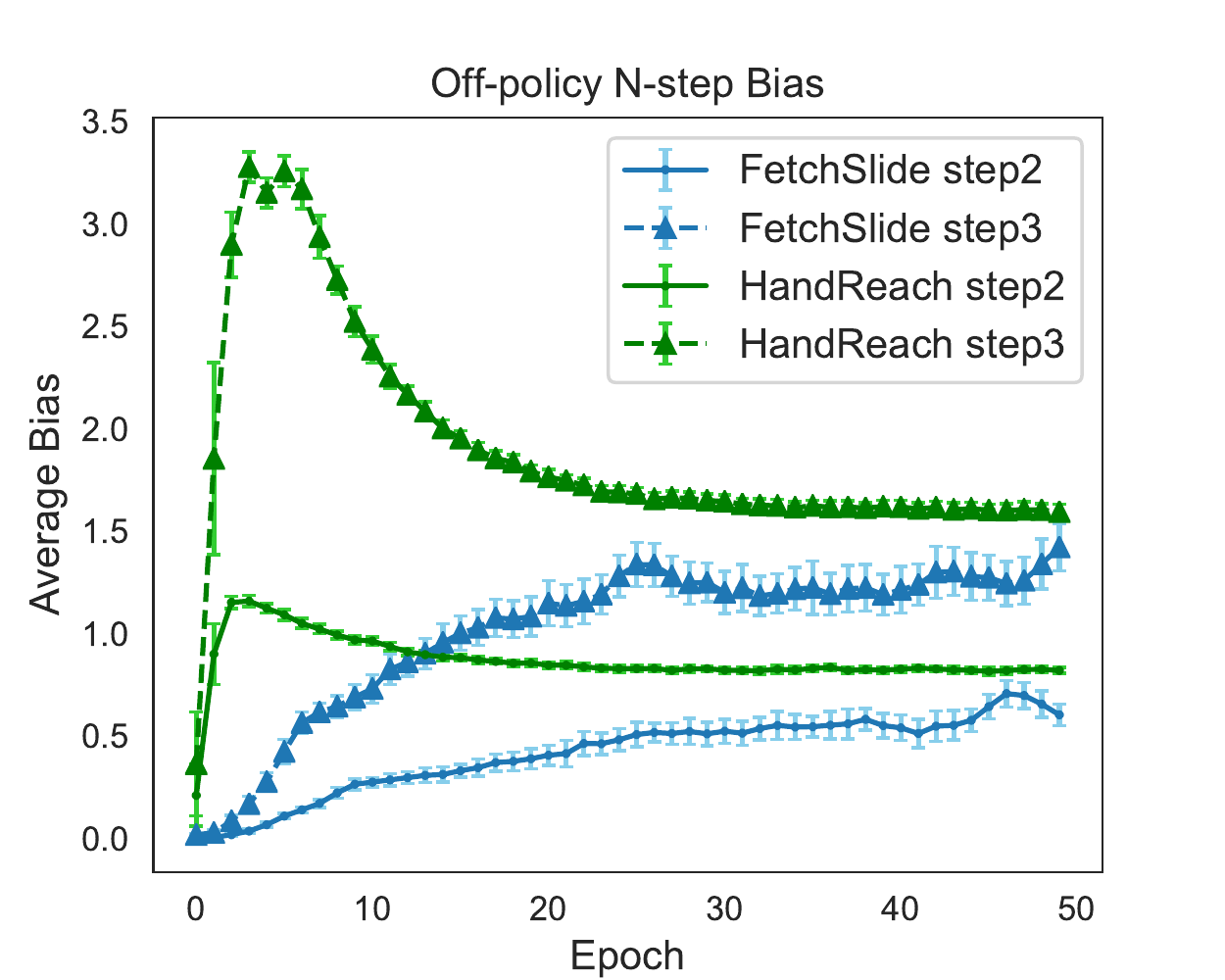}
    \end{minipage}%
}%
\subfigure[]{
    \begin{minipage}[t]{0.45\linewidth}
    \centering
    \includegraphics[width=0.85\linewidth]{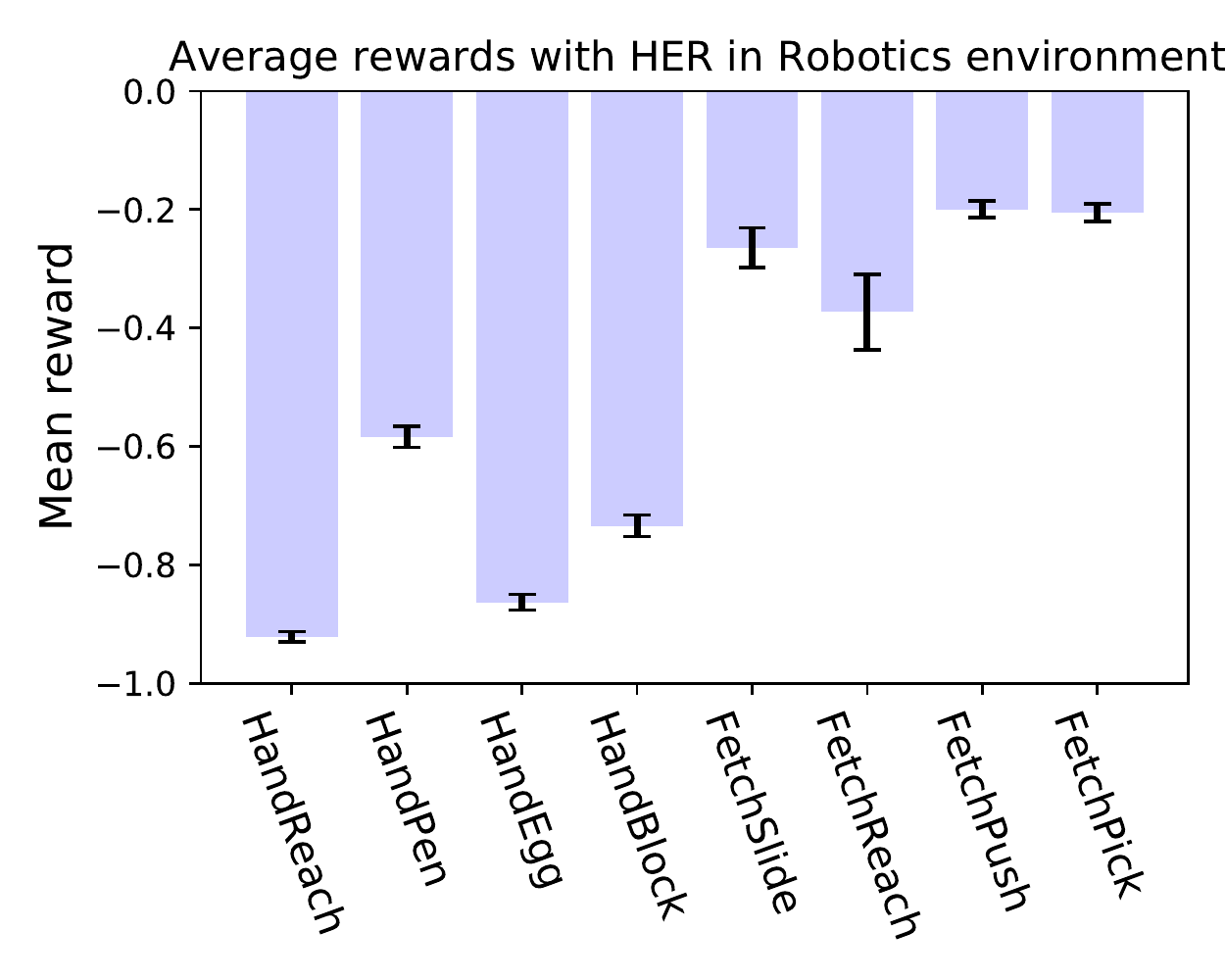}
    \end{minipage}%
}%
\centering
 \caption{(a) Off-policy $n$-step bias training with vanilla MHER in FetchSlide and HandReach environments. (b) Average rewards training using HER in Robotics environments. The rewards are recorded during training HER for $50$ epochs.}
\label{fig:bias_rewards}
\end{figure}

The above analysis is experimentally verified. Average off-policy $n$-step bias computed by Eq. (\ref{eq:off-policybias}) and average reward using HER are plotted in Figure \ref{fig:bias_rewards} (a) and Figure \ref{fig:bias_rewards} (b), respectively. In Figure \ref{fig:bias_rewards} (a), the bias of $3$-step MHER in the same environment is usually larger than $2$-step. Meanwhile, larger absolute average reward in HandReach (see Figure \ref{fig:bias_rewards} (b)) leads to larger bias than FetchSlide. It needs to be emphasized that the bias in Figure \ref{fig:bias_rewards} (a) is estimated using the learned value function rather than the true value function.

\begin{algorithm2e}[htb]
\caption{Functions of Proposed Methods
\label{ag:allalgos}}
\begin{multicols}{2}
\DontPrintSemicolon
 \SetKwFunction{Target}{DefaultTarget}
  \SetKwFunction{Select}{Select-MHER($\lambda$)}
  \SetKwFunction{FMHERlambda}{Target-MHER($\lambda$)}
  \SetKwFunction{FModelSelect}{Select-MMHER}
  \SetKwFunction{FModelbased}{Target-MMHER}
%  \;
 \SetKwProg{Fn}{function}{:}{}
 
%   \Fn{\Target{$*args$}}{
%         Compute $y_t^{(n)}$ according to Eq. (\ref{eq:nsteptarget}) \;
%   }
%   \KwRet $y_t^{(n)}$\;
%   \;
  
  \Fn{\Select{$*args$}}{
        Get trajectory $\mathcal{T}$ and the transition index $t$ \;
        Set $n=min(n, length(\mathcal{T}) - t)$\;
  }
  \KwRet $\{(s_{t+i}, a_{t+i},r_{t+i},s_{t+1+i},g_{t+i}) \}_{i=0}^{n-1}$\;
  \;

  \;

  \SetKwProg{Fn}{function}{:}{}
  \Fn{\FMHERlambda{$*args$}}{
        \For{$i=1,\ldots, n$}{
            Compute $y_t^{(i)}$ using Eq. (\ref{eq:nsteptarget})\;
        }
		$\hat y^{(n)} = \frac{\Sigma_{i=1}^{n} \lambda ^i \cdot y_{t}^{(i)}}{\Sigma_{i=1}^{n} \lambda ^i}$ according to Eq. (\ref{eq:MHERlambda})\;
  }
  \KwRet $\hat y^{(n)}$\;
  \;
  
  \SetKwProg{Fn}{function}{:}{}
  \Fn{\FModelSelect{$*args$}}{
    Get $(s_t,a_t,r_t,s_{t+1},g')$ from $*args$\;
    $b = \{(s_t,a_t,r_t,s_{t+1},g')\}$ \;
    \For{$i=1, \ldots, n-1$}{
        $a_{t+i} = \pi(s_{t+i},g')$\;
        $s_{t+i+1}=model(s_{t+i},a_{t+i})$\;
        $r_{t+i}=reward(s_{t+i+1},g')$\;
        Add $(s_{t+i},a_{t+i},r_{t+i},s_{t+i+1},g')$ to $b$\;
    }
  }
  \KwRet $b$\;
  \;
  
  \SetKwProg{Fn}{function}{:}{}
  \Fn{\FModelbased{$*args$}}{
        Compute $y_t^{(n)}$ according to Eq. (\ref{eq:modeltarget}) \;
		Compute $\hat y^{(n)}$ according to Eq. (\ref{eq:modelweight})\;
  }
  \KwRet $\hat y^{(n)}$\;
  \;
\end{multicols}
\end{algorithm2e}

\section{Bias-reduced MHER Algorithms}
In this section, we present two algorithms to reduce the off-policy $n$-step bias in MHER with the idea of $\lambda$ return \citep{seijen2014true} and model-based $n$-step value expansion \citep{feinberg2018model}. All the functions of proposed methods are shown in Algorithm \ref{ag:allalgos}.

\subsection{MHER($\lambda$)}
Inspired by TD($\lambda$) where a bias-variance trade-off is made between TD($0$) and MC, we propose MHER($\lambda$) to balance between lower bias as one-step target and more reward information as $n$-step target. An exponential decay weight parameter $\lambda\in[0,1]$ is introduced to combine $n$-step targets $y_t^{(i)}, i\in[1,n]$ computed by Eq. (\ref{eq:nsteptarget}):
\begin{equation}
\label{eq:MHERlambda}
 \hat y ^{(n)} = \frac{\sum_{i=1}^{n} \lambda^i y_{t}^{(i)}}{\sum_{i=1}^{n} \lambda^i}.
\end{equation}

For an intuitive analysis on Eq. (\ref{eq:MHERlambda}), $\hat y^{(n)}$ becomes close to $1$-step target without bias if $\lambda \rightarrow 0$ and assigns more weight on $n$-step target as $\lambda$ increases. Therefore, $\lambda$ makes a trade-off between bias and reward information provided by off-policy $n$-step returns. By adjusting the parameter $\lambda$, MHER($\lambda$) can adapt to different environments regardless of environmental difficulties and absolute average rewards.

\subsection{Model-based MHER(MMHER)}
\label{sec:MMHER}
MMHER generates on-policy value expansion with current policy and a learned dynamics model to alleviate the off-policy bias. As discussed in Section \ref{sec:intutive}, on-policy returns don't contain off-policy bias. In MMHER, a model $m:\mathcal{S}\times \mathcal{A}\rightarrow \mathcal{S}$ is trained to fit environmental dynamics by minimizing the following loss:
\begin{equation*}
\label{eq:dynamicloss}
    \mathcal{L}_{dynamics} =E_{(s_t,a_t,s_{t+1})\sim B} \|s_{t+1} - m(s_t,a_t)\|_2^2,
\end{equation*}
where $B$ refers to the replay buffer. Details of training the dynamics model $m$ are presented in Appendix \ref{detailMMHER}.

Fully on-policy experiences cannot benefit from hindsight relabeling, thus we start model-based interaction with the relabeled goal $g'$. Specifically, given a transition in the replay buffer $(s_t,a_t,r_t,s_{t+1},g)$, we relabel it with $g'$ and $r_t'$ according to Section \ref{sec:her}. Then we start from $s_{t+1}$ and use the trained model $m$ and current policy $\pi$ to generate $n-1$ transitions $(s'_{t+i}, a'_{t+i}, r'_{t+i},s'_{t+i+1},g'),i\in [1,n-1], s'_{t+1}=s_{t+1}$. Finally, we expand the target to $n$-step as:
\begin{equation}
\label{eq:modeltarget}
    y_t^{(n)} = \sum_{i=0}^{n-1} \gamma^i r'_{t+i} + \gamma^n Q(s'_{t+n},\pi(s'_{t+n},g'),g'),
\end{equation}
where $r'_t$ can directly benefit from hindsight relabeling. With the relabeled goal $g'$, the expanded target also contributes to the value estimation in sparse reward setting. If the model is accurate, $y_t^{(n)}$ is an unbiased $n$-step estimation of $Q(s_t,a_t,g')$.
% and the other items can be viewed as an on-policy estimation of $\gamma Q(s_{t+1}, \pi(s_{t+1}, g'),g')$. 

However, model-based methods encounter with \emph{model bias} caused by the difference between the trained model and real environment \citep{janner2019trust}, especially when the environment has high-dimensional states and complex dynamics. To balance between model bias and learning information, we utilize a weighted average of $n$-step target and one-step target using a parameter $\alpha$:
\begin{equation}
\label{eq:modelweight}
    \hat y^{(n)}  = \frac{\alpha \times y_t^{(n)} + (r'_t + \gamma Q(s_{t+1},\pi(s_{t+1},g'),g'))}{\alpha + 1}.
\end{equation}
When $\alpha \rightarrow 0$, $\hat y^{(n)}$ is close to unbiased $1$-step return. As $\alpha$ increases, $\hat y^{(n)}$ gains more information from model-based $n$-step returns but meanwhile contains more model bias.

\begin{figure}[htb]
\centering
\subfigure[]{
    \begin{minipage}[t]{0.2\linewidth}
    \centering
    \includegraphics[height=1.6cm, width=2.3cm]{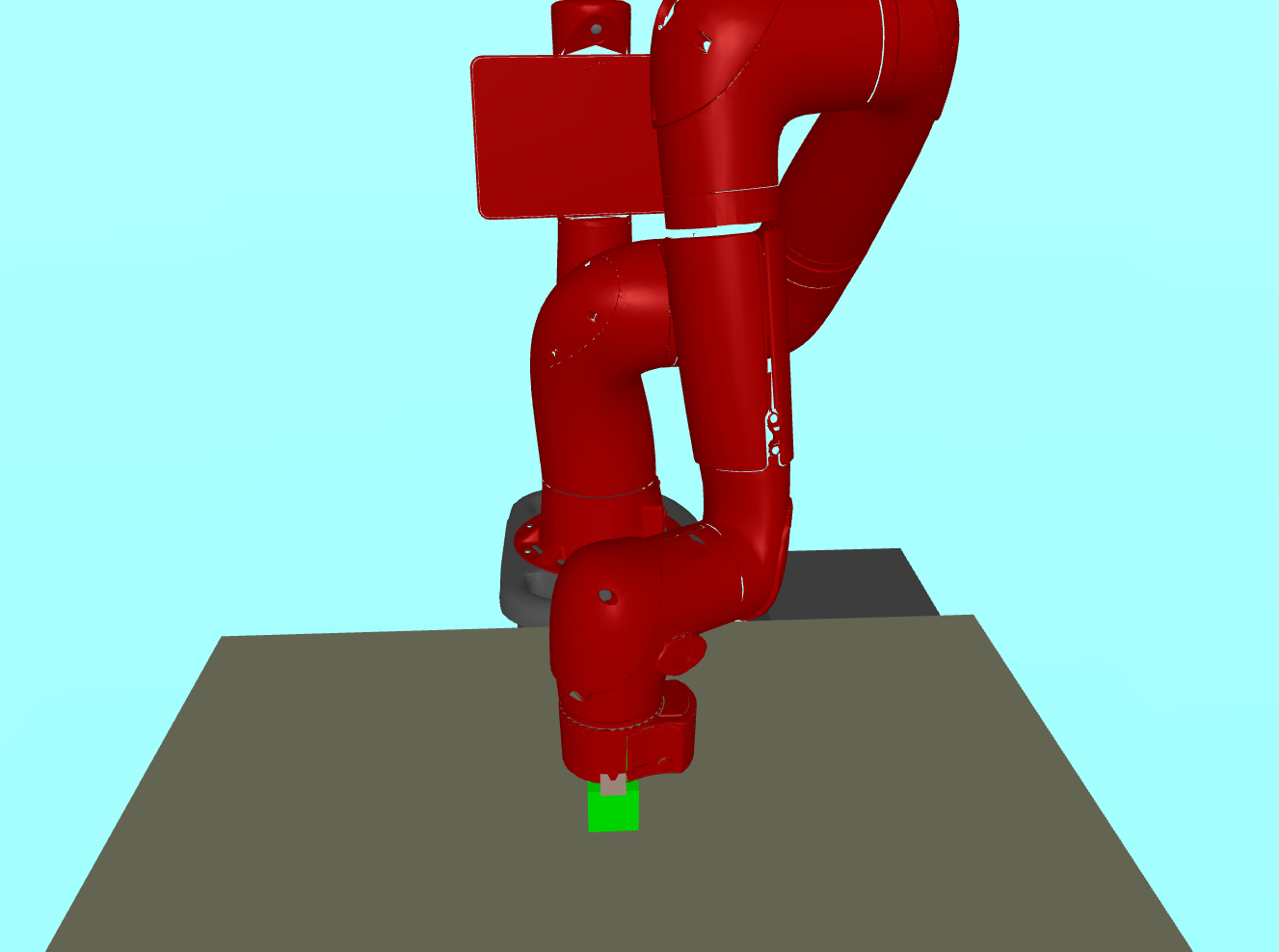}
    \end{minipage}%
}%
\subfigure[]{
    \begin{minipage}[t]{0.2\linewidth}
        \centering
        \includegraphics[height=1.6cm, width=2.3cm]{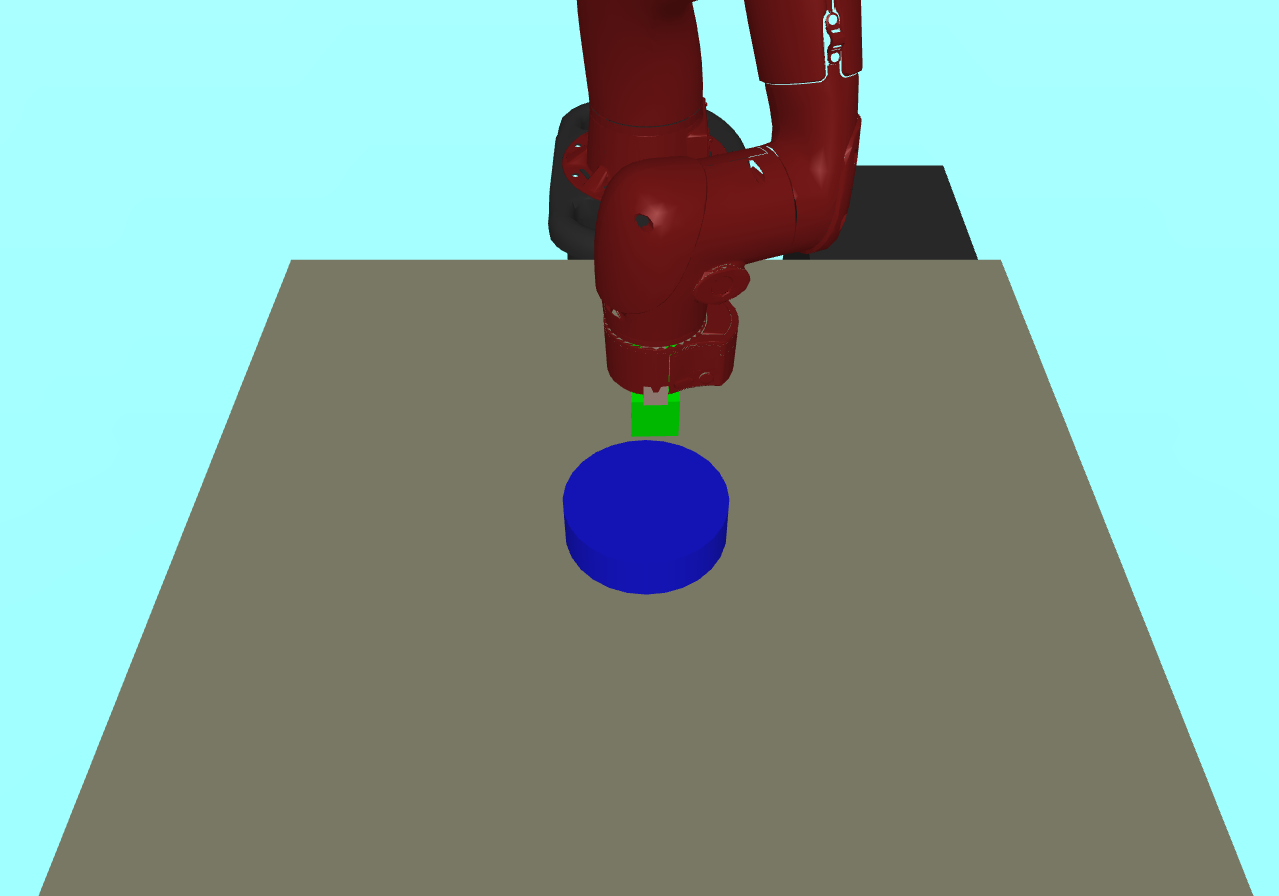}\\
    \end{minipage}%
}%
\subfigure[]{
    \begin{minipage}[t]{0.2\linewidth}
        \centering
        \includegraphics[height=1.6cm, width=2.3cm]{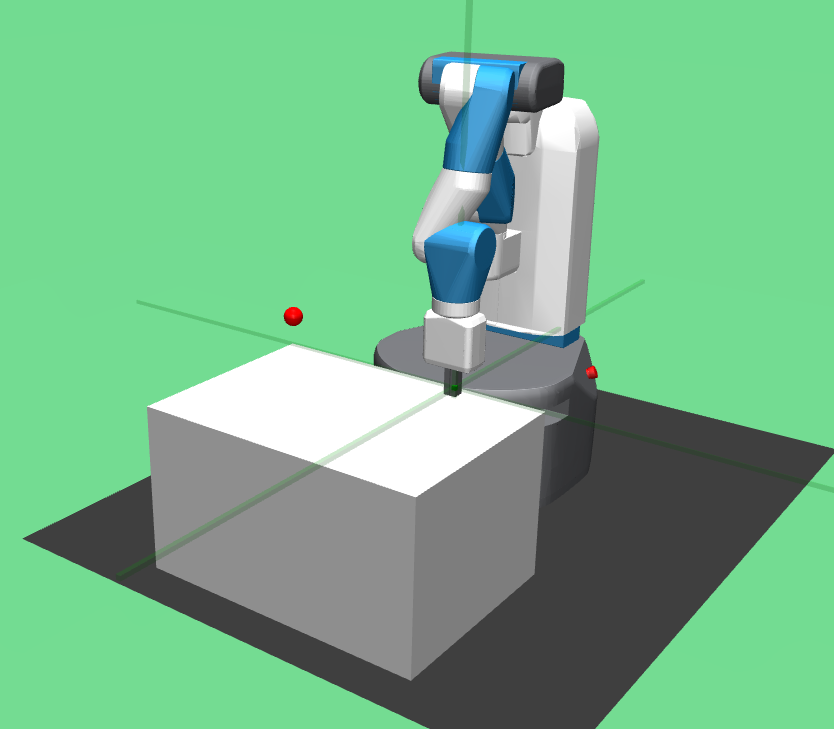}\\
    \end{minipage}%
}%
\subfigure[]{
    \begin{minipage}[t]{0.2\linewidth}
        \centering
        \includegraphics[height=1.6cm, width=2.3cm]{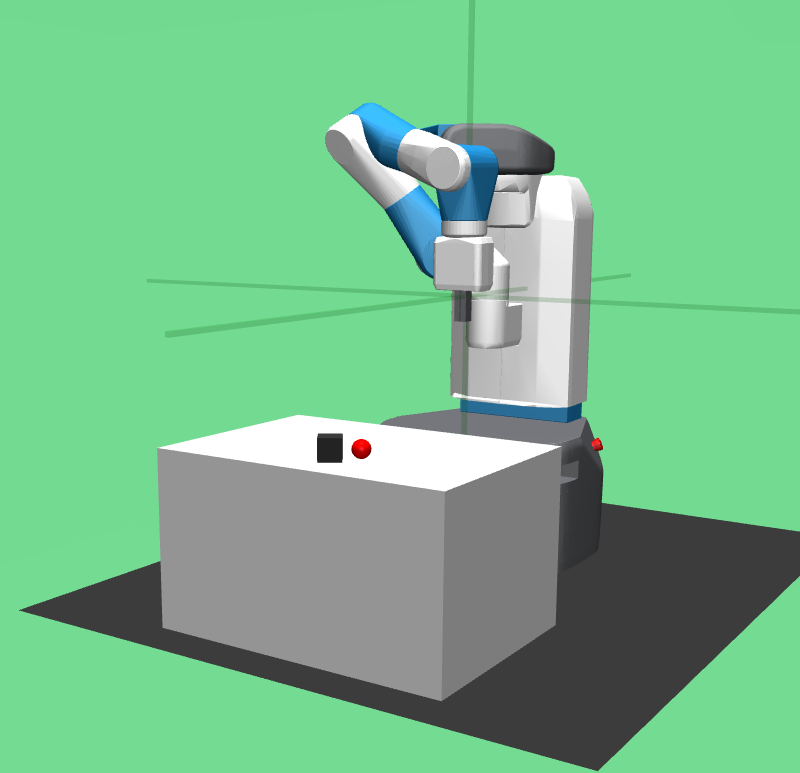}\\
    \end{minipage}%
}%

\subfigure[]{
    \begin{minipage}[t]{0.2\linewidth}
        \centering
        \includegraphics[height=1.6cm, width=2.3cm]{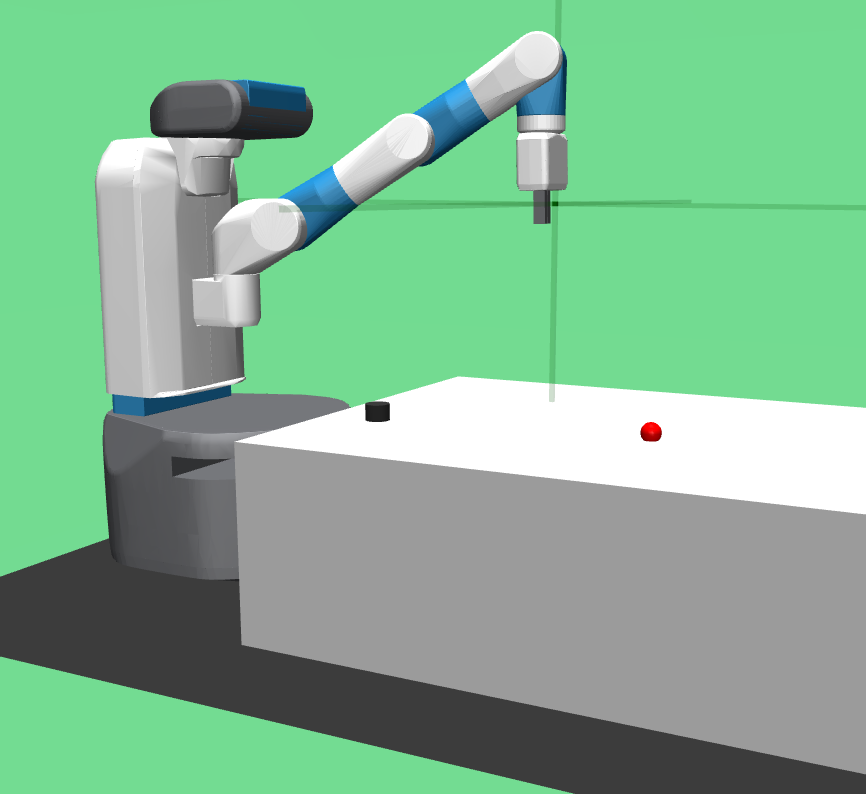}\\
    \end{minipage}%
}%
\subfigure[]{
    \begin{minipage}[t]{0.2\linewidth}
        \centering
        \includegraphics[height=1.6cm, width=2.3cm]{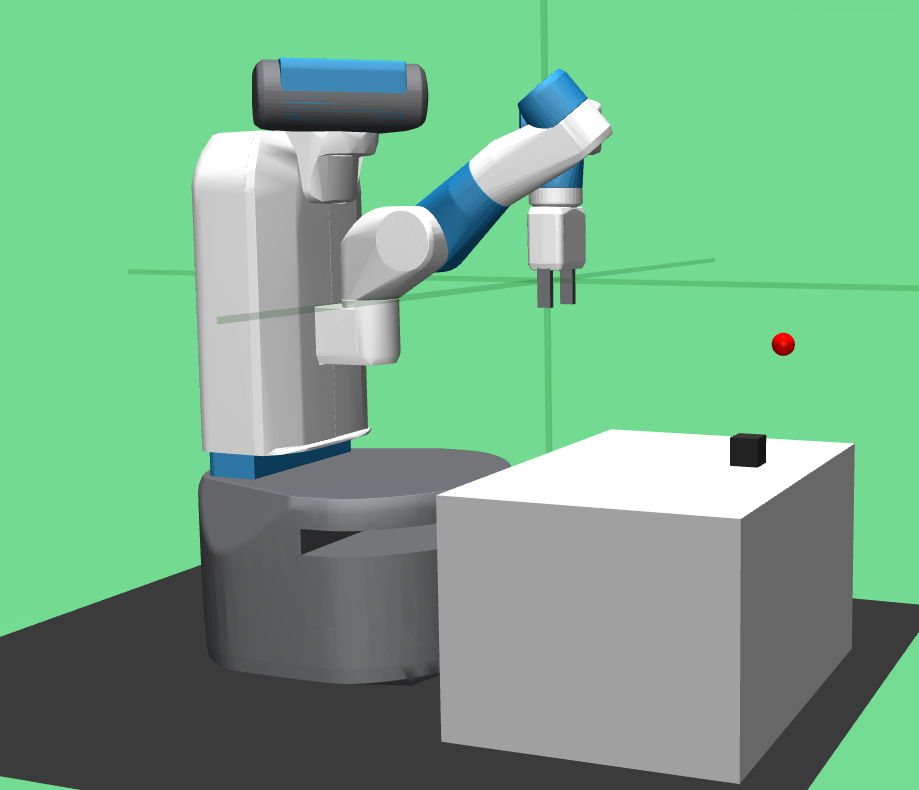}\\
    \end{minipage}%
}%
\subfigure[]{
    \begin{minipage}[t]{0.2\linewidth}
        \centering
        \includegraphics[height=1.6cm, width=2.3cm]{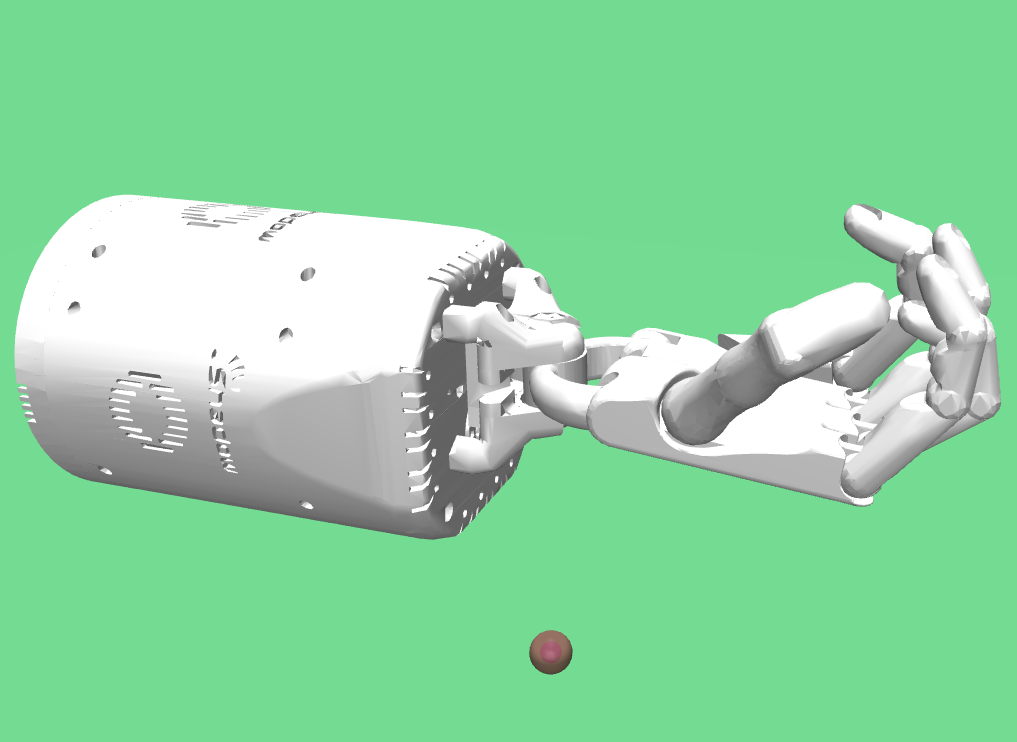}\\
    \end{minipage}%
}%
\subfigure[]{
    \begin{minipage}[t]{0.2\linewidth}
        \centering
        \includegraphics[height=1.6cm, width=2.3cm]{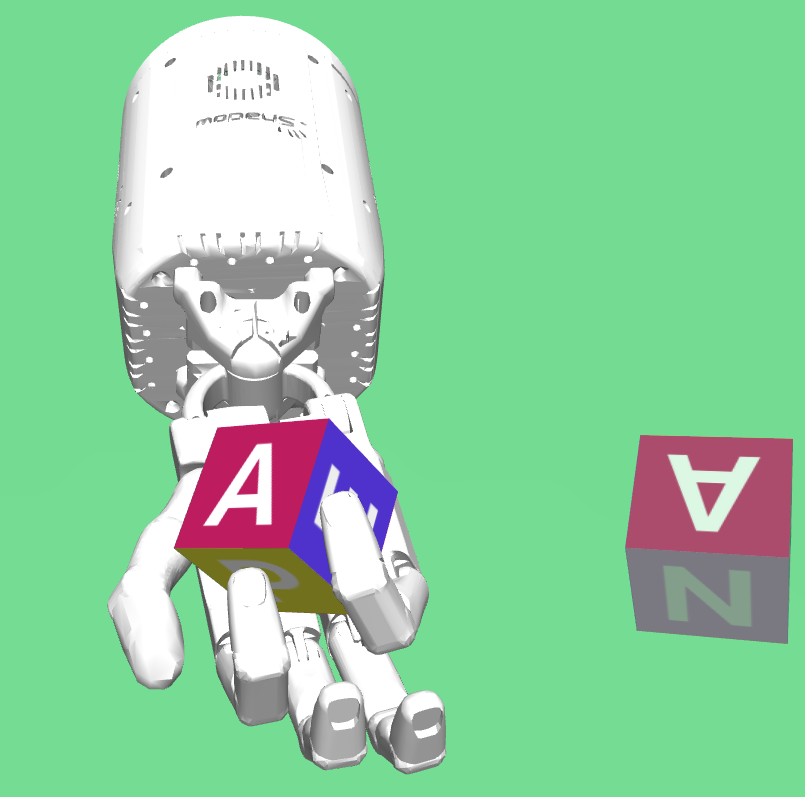}\\
    \end{minipage}%
}%
\centering
\caption{Eight simulated robotics environments: (a) SawyerReachXYZEnv-v1, (b) SawyerPushAndReachEnvEasy-v0, (c) FetchReach-v1, (d) FetchPush-v1, (e) FetchSlide-v1, (f) FetchPickAndPlace-v1, (g) HandReach-v0, (h) HandManipulateBlockRotateXYZ-v0.}
% \vspace{1cm}
% \hspace{0.5cm}
\label{fig:envs}
\end{figure}

\begin{figure}[htb]
\setlength{\abovedisplayskip}{3pt} 
\includegraphics[width=1\linewidth,trim=0 40 0 0]{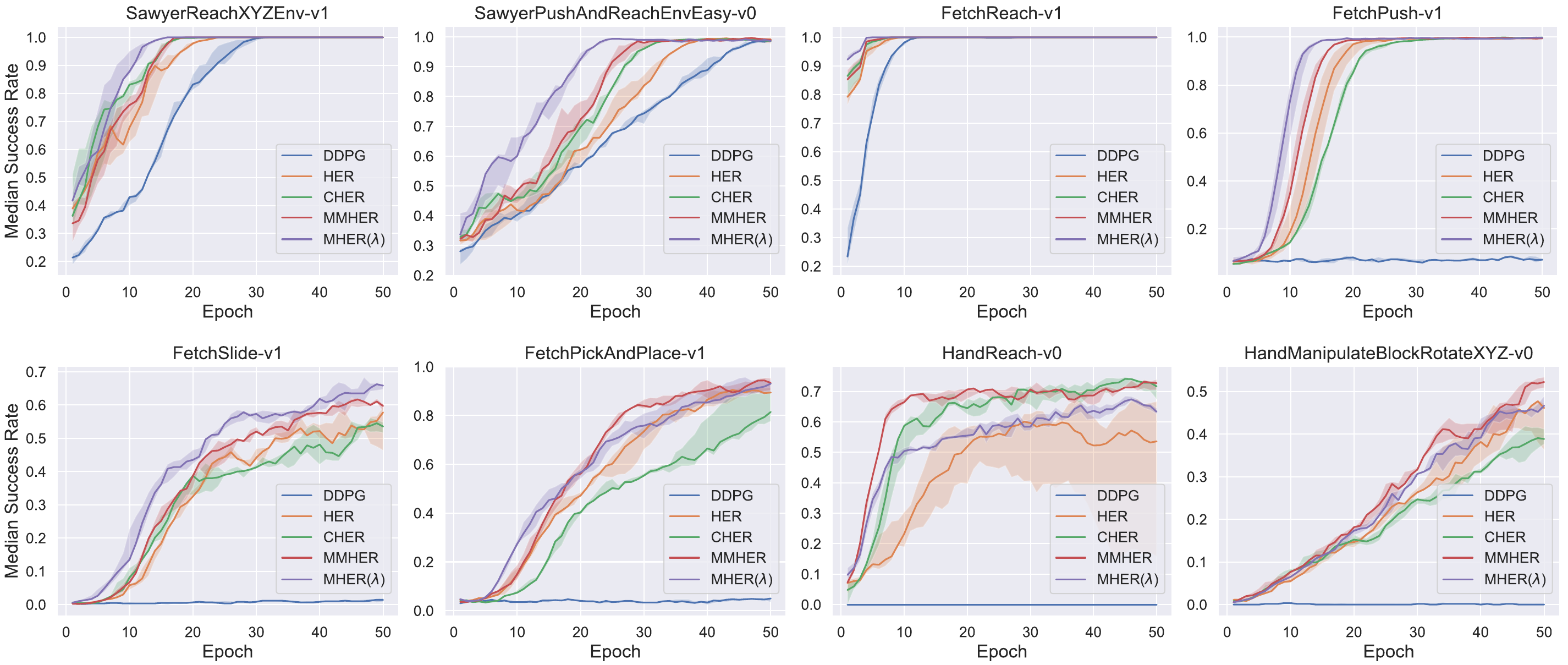}
\caption{Median test success rate (line) with interquartile range (shaded area) in robotics environments.}
\label{fig:allresult}
\end{figure}

\section{Experiments}
We evaluate MHER($\lambda$) and MMHER in eight challenging multi-goal robotics environments and compare them to three baselines: DDPG, original HER, and CHER \citep{fang2019curriculum}. For each experiment, we train for 50 epochs with 12 parallel environments and a central learner, which has the advantage of better utilization of GPU. We sample 12 trajectories in every cycle and train all the algorithms for 40 batches. Three relatively easy tasks (SawyerReach, SawyerPush, and FetchReach), contain 10 cycles each epoch, while other tasks contain 50 cycles each epoch. In order to emphasize the sample efficiency, we only sample $12 \times 50$ trajectories ($12 \times 10$ for the three easy tasks) each epoch and train the central learner with a batch size of $1024$, which is approximately $1/3$ of samples and $1/5$ of the computation in previous work \citep{plappert2018multi}. Detailed hyperparameters are listed in Appendix \ref{ap:hyperpara}. 

In benchmark experiments, we select $n=3,\lambda=0.7$ for MHER($\lambda$) in \emph{Sawyer} and \emph{Fetch} environments, and $n=2, \lambda=0.7$ in \emph{Hand} environments. As for MMHER, the parameter $\alpha$ and step number $n$ are set to $0.4$ and $2$ respectively across all environments.

After training for one epoch, we test without action noise for 120 episodes to accurately evaluate the performance of algorithms. For all the experiments, we repeat experiments using $5$ random seeds and depict the median test success rate with the interquartile range.

\subsection{Environments}
As shown in Figure \ref{fig:envs}, eight challenging robotics environments are included in our experiments. In each environment, agents are required either to reach the desired state or manipulate a specific object to the target pose. Agents receive a reward of $0$ if successfully achieve the desired goal, and a reward of $-1$ otherwise. The \emph{Sawyer}, \emph{Fetch} and \emph{Hand} environments are taken from \citep{yu2020meta} and \citep{plappert2018multi} respectively. A more detailed introduction of environments is as follows.

\subsubsection{Sawyer Environments}The Sawyer robot's observations are $3$-dimensional vectors representing the 3D Cartesian positions of the end-effector. Similarly, goals are $3$-dimensional vectors describing the position of the target place. The action space is also $3$-dimensional which indicates the next expected position of the end-effector.

\subsubsection{Fetch Environments}The robot in Fetch environments is a 7-DoF robotic arm with a two-finger gripper and it aims to touch the desired position or push, slide, place an object to the target place. Observations contain the gripper's position and linear velocities (10 dimensions). If the object exists, another 15 dimensions about its position, rotation, and velocities are also included. Actions are 4-dimensional vectors representing grippers' movements and their opening and closing. Goals are 3-dimensional vectors describing the expected positions of the gripper or the object.

\subsubsection{Hand Environments}Hand environments are constructed with a $24$-DoF anthropomorphic robotic hand and aim to control its fingers to reach the target place or manipulate a specific object (e.g., a block) to the desired position. Observations contain two $24$-dimensional vectors about positions and velocities of the joints. Additional $7$ or $15$ dimensions indicating current state of the object or fingertips are also included. Actions are 20-dimensional vectors and control the non-coupled joints of the hand. Goals have $15$ dimensions for HandReach containing the target Cartesian position of each fingertip or $7$ dimensions for HandBlock representing the object's desired position and rotation.

\begin{figure}[htb]
\centering
\subfigure[]{
    \label{fig:Q_value}
    \begin{minipage}[t]{0.45\linewidth}
        \centering
        \includegraphics[width=0.8\linewidth]{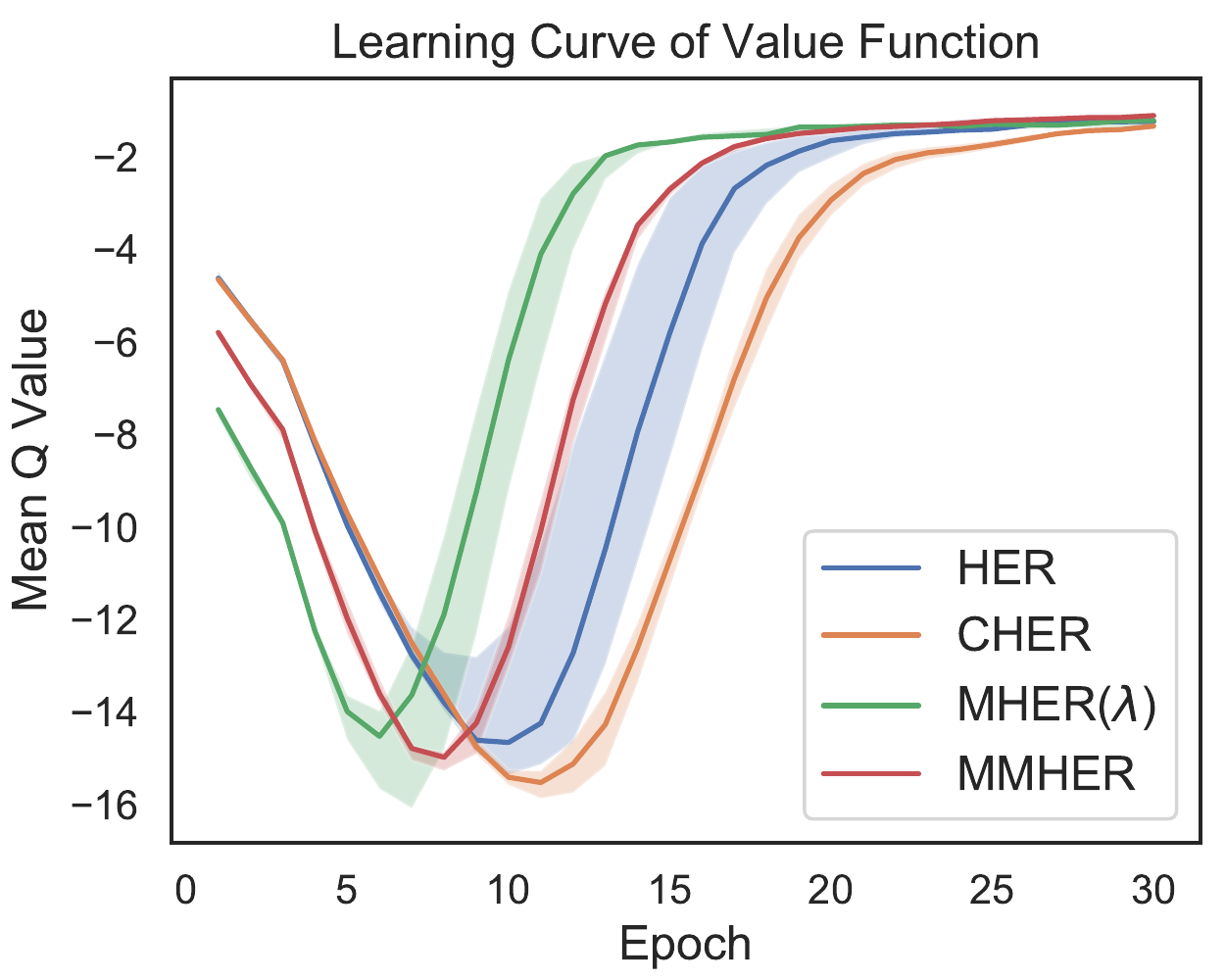}\\
    \end{minipage}%
}%
\subfigure[]{
    \label{fig:time}
    \begin{minipage}[t]{0.45\linewidth}
        \centering
        \includegraphics[width=0.8\linewidth]{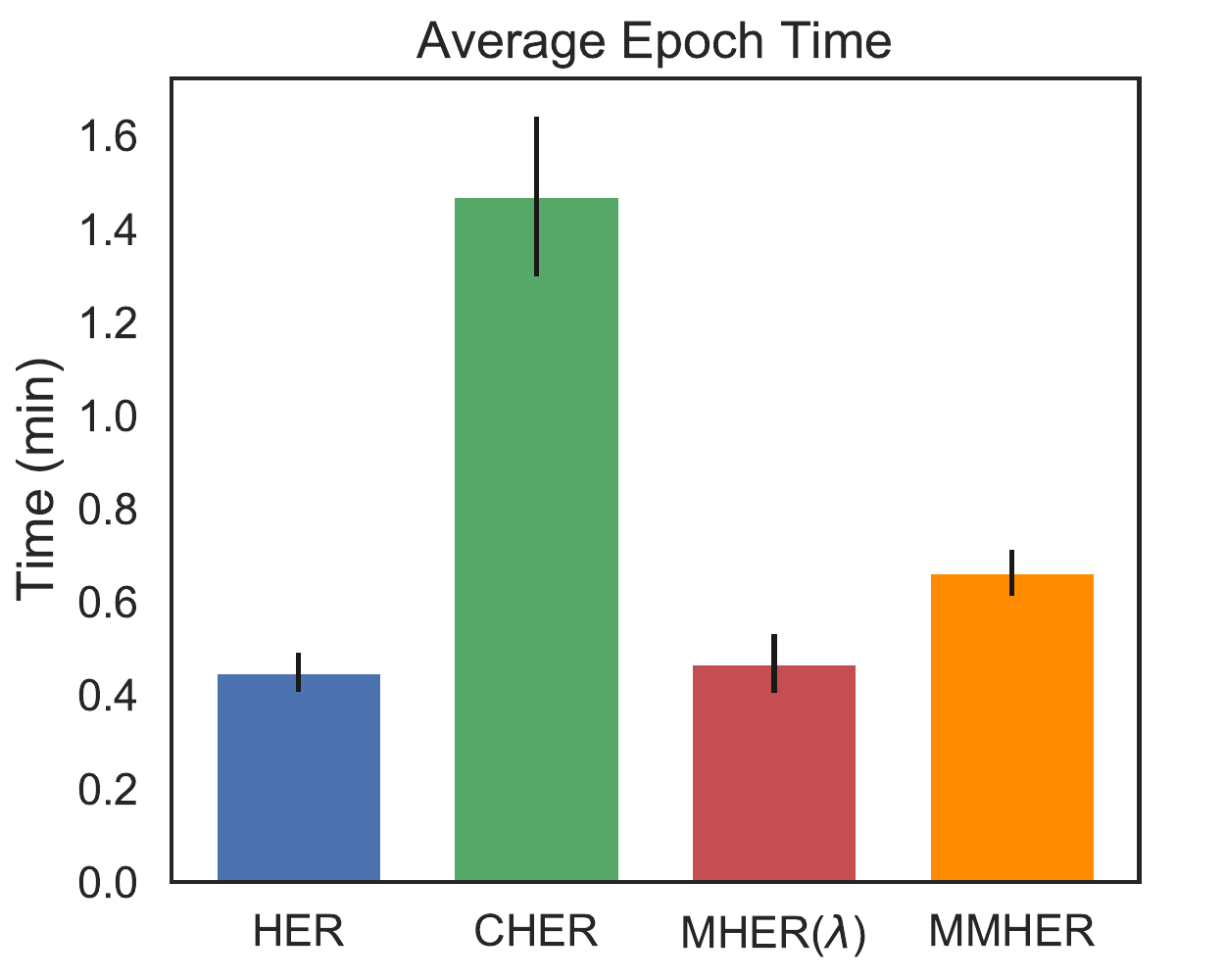}\\
    \end{minipage}%
}%
\centering
\caption{(a) Learning curves of Q value using different algorithms in FetchPush environment, (b) average epoch time of different algorithms in SawyerPushAndReachEnvEasy environment.}
\label{fig:Qandtime}
\end{figure}

\subsection{Benchmark Results}
Figure \ref{fig:allresult} reports the performance of five algorithms in eight environments. The experimental results clearly show that our methods can achieve faster training speed and higher average success rate compared with HER and CHER, even in Hand environments where off-policy $n$-step bias is quite large. Comparing with the results in figure \ref{fig:nstepher}, MHER($\lambda$) and MMHER successfully alleviate the impact of off-policy bias in vanilla MHER and bring considerable performance improvement.
% Besides, MHER($\lambda$) outperforms MMHER in most of the environments except the two Hand environments, which is mainly because MMHER is less affected by off-policy $n$-step bias. 

Furthermore, we depict the learning curves of Q-function and the average epoch time in Figure \ref{fig:Qandtime}. Figure \ref{fig:Qandtime} (a) implies that MHER($\lambda$) and MMHER contribute to fast learning of value functions compared to HER and CHER. On average, the learned value function does not deviate from the convergent value of HER. Figure \ref{fig:Qandtime} (b) provides evidence that our two algorithms also have computational advantages over CHER. The results also demonstrate that MHER($\lambda$) improves performance significantly at less cost of computation resources than CHER and MMHER.

\subsection{MHER($\lambda$) vs MMHER}
In Figure \ref{fig:allresult}, the results demonstrate that MHER($\lambda$) performs better than MMHER in most Sawyer and Fetch environments while MMHER performs better in Hand environments. Considering the difference of average rewards discussed in Figure \ref{fig:bias_rewards} and Section \ref{sec:bias}, we can conclude that MHER($\lambda$) is more considerable when handling tasks with relatively smaller absolute average rewards. On the contrary, when in environments with large absolute average rewards where off-policy $n$-step bias is aggravated, MMHER is more favorable as it utilizes model-based on-policy returns and is less affected by the off-policy bias.

\begin{figure}[htb]
\centering
\subfigure[]{
    \label{fig:parameter_a}
    \begin{minipage}[t]{0.45\linewidth}
        \centering
        \includegraphics[width=0.8\linewidth]{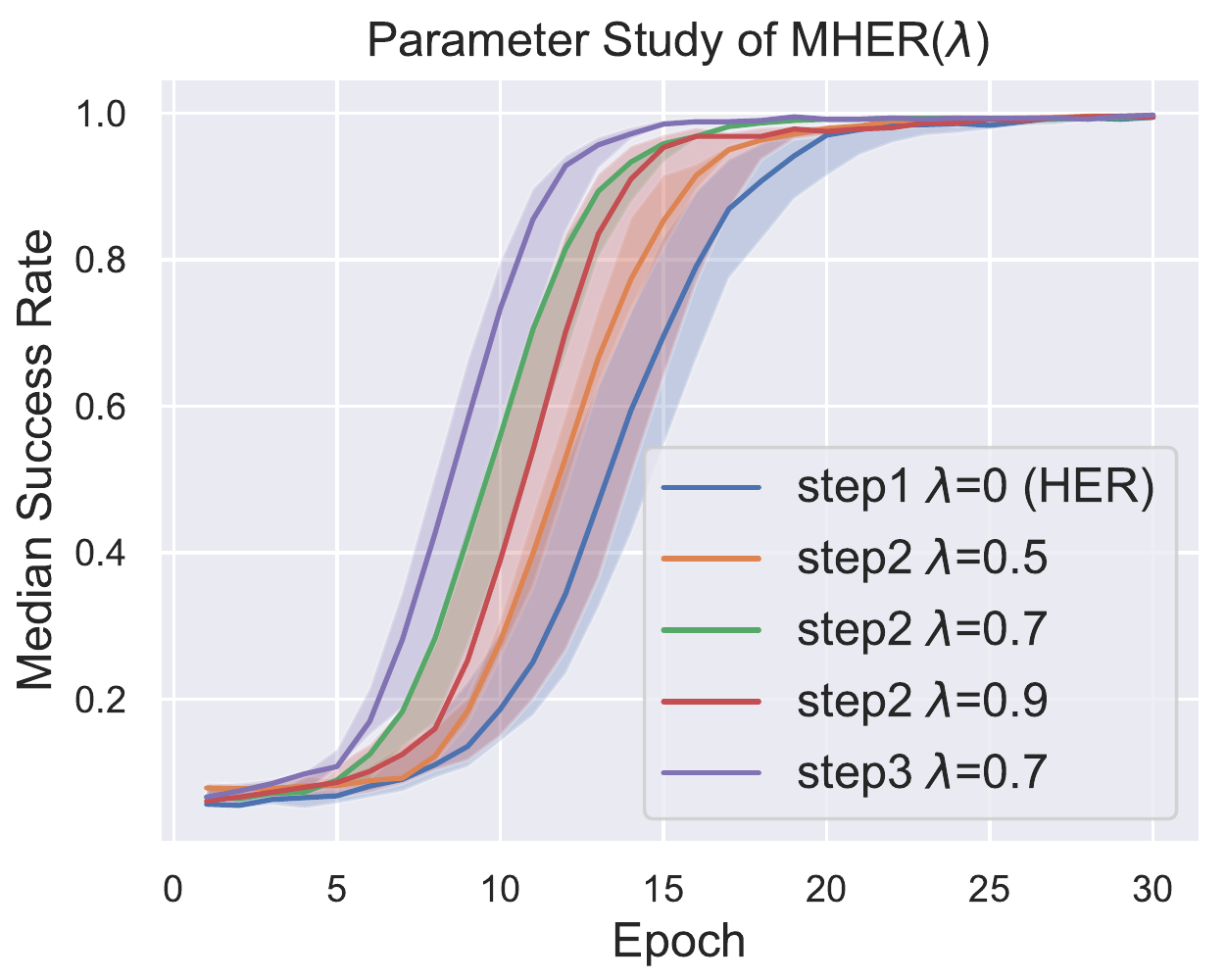}\\
    \end{minipage}%
}%
\subfigure[]{
    \label{fig:parameter_b}
    \begin{minipage}[t]{0.45\linewidth}
        \centering
        \includegraphics[width=0.8\linewidth]{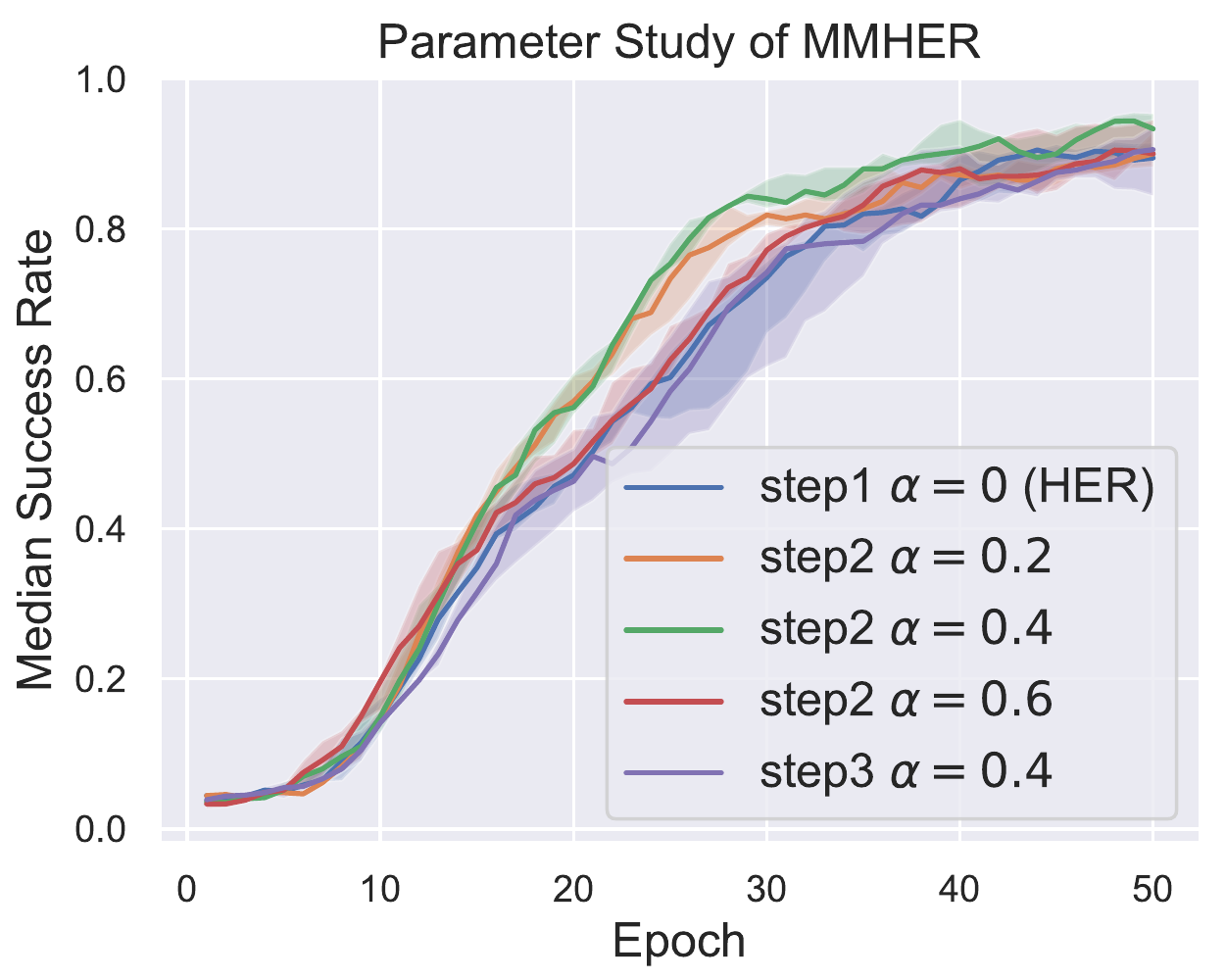}\\
    \end{minipage}%
}%
\centering
\caption{(a) Parameter study for MHER($\lambda$) in FetchPush, (b) parameter study for MMHER in FetchPickAndPlace. }
\label{fig:parameterstudy}
\end{figure}

\subsection{Parameter Study}
We then take an insight into how different parameters affect the performance of MHER($\lambda$) and MMHER. Comparison results in FetchPush and FetchPickAndPlace environments are shown in Figure \ref{fig:parameterstudy}. From Figure \ref{fig:parameterstudy} (a) we can conclude that $\lambda = 0.7$ successfully makes a trade-off between bias and learning speed, i.e., bigger $\lambda$ suffers from off-policy bias and smaller $\lambda$ learns slowly. Moreover, $3$-step MHER($\lambda$) outperforms $2$-step in FetchPush environment. As for MMHER in Figure \ref{fig:parameterstudy} (b), there is also an apparent trade-off between model bias and learning speed with different $\alpha$. The result of $\alpha=0.4$ is the best in $\{0.2, 0.4, 0.6\}$. Besides, 3-step MMHER performs relatively worse than 2-step, which is mainly because the model bias accumulates as the step number $n$ increases.

\section{Conclusion}
Based on our idea of Multi-step Hindsight Experience Replay and analysis about the off-policy $n$-step bias, we propose two effective algorithms, MHER($\lambda$) and MMHER, both of which successfully mitigate the off-policy $n$-step bias and improve sample efficiency in sparse-reward multi-goal tasks. The two algorithms alleviate the off-policy $n$-step bias with an exponentially decreasing weighted sum of $n$-step targets and model-based value expansion, respectively. Experiments conducted in eight challenging robotics environments demonstrate that our two algorithms outperform HER and CHER significantly in sample efficiency at the cost of little additional computation beyond HER. Further analysis of experimental results suggests MHER($\lambda$) is more suitable for environments with small absolute average rewards, while MMHER is more favorable in environments with large absolute average rewards where off-policy $n$-step bias is aggravated.

% \acks{Acknowledgements should go at the end, before appendices and references.}

%\bibliographystyle{plain}
\bibliography{main}

\appendix

\section{Proof of Proposition 1}
\label{ap:proof1}
% This section provides an analysis of off-policy $n$-step bias. 
\proof For deterministic policy, action-value function $Q(s_t,a_t)$ has the following unbiased estimation:
\begin{equation*}
\label{eq:Q}
\begin{split}
    Q(s_t,a_t) &= \mathbb{E}_{r_t \sim r(s_t,a_t),s_{t+1} \sim p(\cdot|s_t,a_t)}\big[r_t + \gamma \mathbb{E}_{\pi}[Q(s_{t+1}, \cdot)]\big] \\
                &= \mathbb{E}_{r_t \sim r(s_t,a_t),s_{t+1} \sim p(\cdot|s_t,a_t)}\big[r_t + \gamma Q(s_{t+1},\pi(s_{t+1}))\big],
\end{split}
\end{equation*}
where $p$ refers to the state transition probability and $r$ refers to reward function. The following equations also hold by recursion:

\begin{equation}
\label{eq:estimationofQ}
\begin{aligned}
        Q(s_t, a_t) &= \mathbb{E}[r_t + \gamma Q(s_{t+1},\pi(s_{t+1}))] \\
        &=  \mathbb{E}[r_t + \gamma Q(s_{t+1}, a_{t+1}) + \gamma Q(s_{t+1},\pi(s_{t+1}))  - \gamma Q(s_{t+1},a_{t+1})]\\
        &= \mathbb{E}[r_t + \gamma \mathbb{E}[r_{t+1} + \gamma Q(s_{t+2},\pi(s_{t+2}))] +  \gamma (Q(s_{t+1},\pi(s_{t+1})) - Q(s_{t+1},a_{t+1}))]\\
        &= \ldots \\
        & = \mathbb{E}  \Bigg [ \sum_{i=0}^{n-1} \gamma^i r_{t+i} + \gamma ^n Q(s_{t+n},\pi(s_{t+n})) +  \sum_{i=1}^{n-1} \gamma^i \big [Q(s_{t+i},\pi(s_{t+i})) -Q(s_{t+i},a_{t+i}) \big] \Bigg].
\end{aligned}
\end{equation}

Eq. (\ref{eq:estimationofQ}) gives an unbiased estimation of $Q(S_t,a_t)$. Comparing Eq. (\ref{eq:estimationofQ}) to the $n$-step target in Eq. (\ref{eq:d4pgtarget}), we conclude that off-policy $n$-step bias $\mathcal{B}_t^{(n)}$ at time $t$ has the following formula:
\begin{equation*}
    \mathcal{B}_t^{(n)} = \sum_{i=1}^{n-1} \gamma^i [Q(s_{t+i},\pi(s_{t+i}))-Q(s_{t+i},a_{t+i})].
\end{equation*}

% \section{Off-policy Correction for Deterministic Policy}
% The most common way to correct policy mismatch is importance sampling (IS), but we show that off-policy correction for deterministic policy is actually one-step return. Given the trajectory under policy $\mu$: $(s_t, a_t, r_t, s_{t+1}), t\in[0, n-1]$, the corrected $n$-step target for policy $\pi$ at time step $t$ is as follows \citep{munos2016safe}:
% \begin{equation}
% \label{eq:offpolicycorrection}
% \begin{aligned}
%     y_t^{(n)} = \sum_{i=0}^{n-1} \gamma^i   \left [\prod_{k=0}^{i-1}\frac{\pi(a_{t+k}|s_{t+k})}{\mu(a_{t+k}|s_{t+k})} \right](r_{t+i} +\gamma \mathbb{E}_{\pi} Q(s_{t+i+1}, \cdot) ) .
% \end{aligned}
% \end{equation}

% Stochastic policy maps a state to an action distribution while
% deterministic policy maps a state to a specific action, which can be expressed as
% $
% \mu(a|s)=\begin{cases}1, & \mu(s)=a
% \cr 0, &\text{otherwise}\end{cases}
% % $.
% Therefore, the IS ratio satisfies:

% \begin{equation*}
% \frac{\pi(a_t|s_t)}{\mu(a_t|s_t)}=\begin{cases}1, & \pi(s_t)=a_t
% \cr 0, &\text{otherwise}\end{cases}.
% \end{equation*}
% In continuous control with deterministic policy, $\mathbb{P}(\pi(s_t)=a_t|\pi \neq \mu)=0$. As a result, the corrected $n$-step target in Eq. (\ref{eq:offpolicycorrection}) for deterministic policy $\pi$ only keeps the first item with a great probability.

% The above proof shows that multi-step rewards are not utilized with off-policy correction. On the contrary, our two algorithms exploit $n$-step returns more fully and safely. 

\section{Proof of Proposition 2}
\label{ap:proof2}
% In this section, we give a proof of Proposition \ref{propsitionaveragebias}. 
\proof The reward function defined in Eq. (\ref{equ:rewardfunction}) limits reward $r_t \in [-1,0]$ and value function $V^{\pi}(s) \leq 0$. Using the definition in Section \ref{sec:bias}, we have 
\begin{equation*}
\begin{aligned}
    \mathcal{B}_t^{(n)} = \sum_{i=1}^{n-1} \gamma^i [Q(s_{t+i},\pi(s_{t+i}))-Q(s_{t+i},a_{t+i})] 
    =\sum_{i=1}^{n-1} \gamma^i [V^{\pi}(s_{t+i})- r_{t+i} - \gamma V^{\pi}(s_{t+i+1})] .
    % &=\sum_{i=1}^{n-1} \gamma^i [-r(s_{t+i},a_{t+i}) + V^{\pi}(s_{t+i}) - \gamma V^{\pi}(s_{t+i+1})]
\end{aligned}
\end{equation*}
In the above formula, $V^{\pi}(s_{t+i})$ and $  V^{\pi}(s_{t+i+1})$ are expected return with current policy, but $r_{t+i}$ is the past reward in the replay buffer. Naturally we will consider grouping terms of $V^{\pi}$ and $r$ separately. In addition, the Lipschitz continuous condition of $V^{\pi}$ holds for any state:
\begin{equation*}
    |V^{\pi}(s) - V^{\pi}(s')| \leq L \|s - s'\|, \forall s,s' \in \mathcal{S}
\end{equation*}
which is a general analysis tool for RL with neural networks \citep{d2019sharing}. 
We then bound average off-policy $n$-step bias $B^{(n)}$ as:
\begin{equation*}
\begin{aligned}
    &\mathcal{B}^{(n)}=\mathbb{E}_{(s_{t+i},a_{t+i}, r_{t+i},s_{t+i+1})_{i=0}^{n-1}\sim B}\big[ \mathcal{B}_t^{(n)} \big] \\
    &= \mathbb{E} \big[\sum_{i=1}^{n-1} \gamma^i [- r_{t+i} + (1-\gamma)V^{\pi}(s_{t+i}) + \gamma(V^{\pi}(s_{t+i})-V^{\pi}(s_{t+i+1}))]  \big] \\
    &\leq \mathbb{E} \big[\sum_{i=1}^{n-1} \gamma^i [- r_{t+i} + \gamma L \|s_{t+i} - s_{t+i+1}\|]\big]\\
    &\leq \gamma(n-1) \mathbb{E}_{(s_t,a_t,r_t,s_{t+1})\sim B}  [- r_t + \gamma L \|s_{t} - s_{t+1}\|] \\
    &= \gamma(n-1)\big[ |\mathbb{E}_{r_t\sim B} [r_t]| + \gamma L E_{(s_t,s_{t+1})\sim B} \|s_t - s_{t+1}\|\big]
\end{aligned}
\end{equation*}
The upper bound is tight when $\gamma$ is close to $1$, $V^{\pi}$ is close to $0$, and the Lipschitz continuous bound is tight.

% The lower bound of $B^{(n)}$ can be obtained by considering the gradient assent of $\mathbb{E}_{s\sim B}[Q(s, \pi(s))]$:
% \begin{equation*}
% \begin{aligned}
%         &\mathcal{B}^{(n)}=\sum_{i=1}^{n-1} \gamma^i\big[\mathbb{E}_{(s,a)\sim B}[Q(s,\pi(s)) - Q(s,a)]\big] \\
%         &= (\sum_{i=1}^{n-1} \gamma^i)\big[\mathbb{E}_{s\sim B}[Q(s,\pi(s))]-\mathbb{E}_{s\sim B}[Q(s,\mu(s))]\big] \geq 0
% \end{aligned}
% \end{equation*}
% where $\mu$ is the data collecting policy of the distribution of $B$. 
% When in MHER setting, $\mu$ is the policy of the relabeled data distribution.

\section{Implementation Details of MMHER}
\label{detailMMHER}
The way of training dynamics model in MMHER has an important impact on the performance because of the complex dynamics of robotic environments. To deal with different scales in multi-dimensional states, we first normalize the input with running mean $\mu$ and standard deviation $\sigma$: $\overline s_t = (s_t - \mu) / \sigma$. Then, we train the dynamics model $m$ (8 layers, 256 neurons each layer) to fit the difference between normalized states $\overline s_t$ and $\overline s_{t+1}$:
\begin{equation*}
    \mathcal{L}'_{dynamic} = \mathbb{E}\|\overline s_{t+1} - \overline s_t - m(\overline s_t,\overline a_t)\|_2^2.
\end{equation*}
Finally, the prediction of next state $s_{t+1}$ is computed by: 
$s_{t+1} = \overline s_{t+1} \times \sigma + \mu$, where $\overline s_{t+1} = \overline s_t + m(\overline s_t, \overline a_t)$. 

As for training details, the learning rate of model $m$ is set to 0.001 and the optimizer is Adam. During the warmup period, we train $m$ for 100 updates with a batch size of $512$. In the formal training phase, we update $m$ 2 times for each training batch.

% how MMHER work
% Our implementation of MMHER not only utilize hindsight relabeling in the first term. Actually, after training with the original experiences, the agent's policy has a higher probability to accomplish the relabeled goals sampling from achieved goals. This also contributes to MMHER's performance.  

\section{Off-policy Correction for Deterministic Policy}
The most common way to correct policy mismatch is importance sampling (IS), but we didn't use it in our paper. The reason is that off-policy correction for deterministic policy is actually one-step return, where multi-step rewards are not exploited. 

Given the trajectory under policy $\mu$: $(s_t, a_t, r_t, s_{t+1}), t\in[0, n-1]$, the corrected $n$-step target for policy $\pi$ at time step $t$ is as follows \citep{munos2016safe}:
\begin{equation}
\label{eq:offpolicycorrection}
\begin{aligned}
    y_t^{(n)} = \sum_{i=0}^{n-1} \gamma^i   \left [\prod_{k=0}^{i-1}\frac{\pi(a_{t+k}|s_{t+k})}{\mu(a_{t+k}|s_{t+k})} \right](r_{t+i} +\gamma \mathbb{E}_{\pi} Q(s_{t+i+1}, \cdot) )
\end{aligned}
\end{equation}
Stochastic policy maps a state to an action distribution while
deterministic policy maps a state to a specific action, which can be expressed as
$
\pi(a|s)=\begin{cases}1, &\pi(s)=a
\cr 0, &\text{otherwise}\end{cases}
$.
Therefore, the IS ratio satisfies:

\begin{equation*}
\frac{\pi(a_t|s_t)}{\mu(a_t|s_t)}=\frac{\pi(a_t|s_t)}{1}=\begin{cases}1, &\pi(s_t)=a_t
\cr 0, &\text{otherwise}\end{cases}.
\end{equation*}
In continuous control with deterministic policy, $\mathbb{P}(\pi(s_t)=a_t|\pi \neq \mu)=0$. As a result, the corrected $n$-step target in Eq. (\ref{eq:offpolicycorrection}) for deterministic policy $\pi$ only keeps the first item with a great probability.

The above proof shows that multi-step rewards are not utilized with off-policy correction. On the contrary, our two algorithms, MHER($\lambda$) and MMHER can exploit $n$-step returns more fully and safely.

\section{Correction With Estimated Bias}
We also tried to correct the $n$-step off-policy bias using $B_t^{(n)}$ in Proposition \ref{propositionbias}. However, we found the performance is not satisfied as MHER($\lambda$) and MMHER. There are two possible reasons: 
\begin{itemize}
    \item[1)] the $Q$ in Proposition \ref{propositionbias} is the true action-value function but we only use an estimated Q function with a neural network, therefore estimated $B_t^{(n)}$ is also biased.
    \item[2)] even assuming $Q$ is unbiased, the variance of $B_t^{(n)}$ can also harm the performance.
\end{itemize}

The implementation code of this method can also be found in our anonymous code link.

\section{Hyperparameters for Experiments}
\label{ap:hyperpara}
The detailed hyperparameters used in this paper are listed in Table \ref{tab:parameter}.

\begin{table}
 \centering
 \caption{Detailed hyperparameters setting}
 \label{tab:parameter}
 \begin{tabular}{p{190pt}p{30pt}}
 \hline
 { \textbf{Hyperparameter} } & {\textbf{value}}\\
 \hline
 $|a_t|_{\mathrm{max}}$ & 1 \\
 Layers & 3 \\
 Number of neurons & 256\\
 Critic learning rate & 0.001 \\
 Actor learning rate & 0.001 \\
 Optimizer & Adam \\
 Buffer size & 1e6 \\
 Polyak averaging coefficient & 0.95 \\
 Quadratic penalty on actions & 1.0 \\
 Observation clip boundary & 200 \\
 Random init episodes  & 100 \\
 Number of cycles per epoch & 50 \\
 Rollout batch size & 1 \\
 Training batches per cycle & 40 \\
 Batch size & 1024 \\
 Number of test rollouts per epoch & 10\\
 Exploration rate & 0.3 \\
 Gaussian action noise std & 0.2 \\
 Replay strategy for HER & future \\
 Number of goals used for relabel & 4 \\
 Observation normalization epsilon & 1e-4 \\
 Normalized observations clip boundary & 5 \\
 Number of steps in MHER & 2/3 \\
 Weight parameter $\lambda$ in MHER($\lambda$) & 0.7\\
 Weight parameter $\alpha$ in MMHER & 0.4\\
 Dynamics model layers & 8\\
 Dynamics model neurons & 256\\
 Learning rate of dynamics model & 1e-3\\
 Batchsize for dynamics model & 512\\
 Warmup update times for MMHER & 100\\
 \hline
 \end{tabular}
\end{table}

\end{document}